\documentclass[final,5p,times,twocolumn]{elsarticle}

\usepackage{natbib}
\usepackage{amsmath,amsfonts}
\usepackage{algorithmic}
\usepackage{algorithm}
\usepackage{tabularx} 
\usepackage{array}
\usepackage{xcolor}   
\newcommand{\zyc}[1]{\textcolor{red}{#1}} 
\usepackage[caption=false,font=normalsize,labelfont=sf,textfont=sf]{subfig}
\usepackage{textcomp}
\usepackage{stfloats}
\usepackage{url}
\usepackage{verbatim}
\usepackage{graphicx}
\usepackage{hyperref}
\usepackage{multirow}
\usepackage{CJKutf8}
\def\x{{\mathbf x}}

\begin{document}

\begin{frontmatter}



\title{FMASH: Advancing Traditional Chinese Medicine Formula Recommendation with Efficient Fusion of Multiscale Associations of Symptoms and Herbs} 


\author[1,2]{Xinhan Zheng\fnref{lab1}} 
\author[1,2]{Xueting Wang\fnref{lab1}} 
\author[1]{Ruotai Li\corref{cor1}\fnref{lab1}}
\ead{rtli@bupt.edu.cn}
\author[3]{Huyu Wu} 
\author[1]{Haopeng Jin} 
\author[1,3]{Yehan Yang}
\author[1]{Guodong Shan} 
\fntext[lab1]{These authors contributed equally to this work.}
\cortext[cor1]{Corresponding author}
\affiliation[1]{organization={Beijing University of Posts and Telecommunications},
            addressline={No.10 Xitucheng Road,Haidian District }, 
            city={Beijing},
            postcode={100876}, 
            state={},
            country={China}}
\affiliation[2]{organization={University of Science and Technology of China},
            addressline={No.100 Fuxing Road}, 
            city={Hefei},
            postcode={230026}, 
            state={Anhui},
            country={China}}
\affiliation[3]{organization={University of Chinese Academy of Sciences},
            addressline={No.1 East Yanqi Lake Road,Huairou District}, 
            city={Beijing},
            postcode={101408}, 
            state={},
            country={China}}

\begin{abstract}
Traditional Chinese medicine (TCM) exhibits remarkable therapeutic efficacy in healthcare through patient-specific formulas. However, current AI-based TCM formula recommendation models and methods mainly focus on data-based textual associations between symptoms and herbs, and have not fully utilized their features and relations at different scales, especially at the molecular scale. To address these limitations, we propose the Fusion of Multiscale Associations of Symptoms and Herbs (FMASH), a novel framework that effectively incorporates the properties of herbs on different scales with clinical symptoms and provides refined embeddings of their multiscale associations. The framework integrates molecular-scale features and macroscopic properties of herbs and combines complex local and global relations in the heterogeneous graph of symptoms and herbs. Moreover, it provides effective representation embeddings of the multiscale features and associations of symptoms and herbs in a unified semantic space. Based on this, the framework is applicable to both the traditional unordered herbal recommendation task and the sequential generative task of herbal formulas. It also improves the performance of the model in both tasks. Comprehensive experiments have been conducted on FMASH, and the results demonstrate that our FMASH-based model outperforms the state-of-the-art (SOTA) model on both datasets, confirming the effectiveness of FMASH in building the TCM formula recommendation model. In Dataset1, our model has achieved a significant improvement compared to the SOTA model, with increases of 3.38\% in Precision@5, 3.89\% in Recall@5, and 3.69\% in F1-score@5. In Dataset2, Precision@5, Recall@5, and F1-score@5 increase by 2.64\%, 1.92\%, and 2.23\%, respectively. This work facilitates the application of the AI-based TCM formula recommendation and promotes the innovative development of TCM diagnosis and treatment. 
\end{abstract}




\end{frontmatter}



\section{Introduction}
Traditional Chinese Medicine (TCM), a profound medical system, plays a significant role in disease treatment and healthcare through the recommendation of patient-specific TCM formulas. The intricate and latent multiscale associations between symptoms and herbs \cite{guo2019deep} result in the complexity of its theoretical system and clinical practice. This leads to the core principle of the "multi-component, multi-target, and multi-approach" treatment of TCM.

Recently, various studies have made some meaningful attempts on the modernization of TCM \cite{zhou2021fordnet,chen2020,li2021,pan2023,wang2019,wang2019tcm,liu2019,zhao2022tcm,jin2023meta}, which have facilitated the scientific recommendation and analysis of the efficacy of TCM formulas and propelled the advancement of TCM. In particular, Ji et al. proposed a multi-content topic model, assuming intrinsic connections between symptoms and herbs \cite{ji2017}. Jin et al. used Graph Convolutional Networks (GCNs) to model relations between symptoms and herbs through collaborative graphs \cite{jin2020}. Ruan et al. employed autoencoders for the embedding of a heterogeneous weighted network that extracts the features of symptoms and herbs more effectively \cite{ruan2019}. Yao et al. introduced topic modeling to explore TCM prescriptions \cite{yao2018}. Yang et al. integrated herb attribute information into a multi-graph convolution model \cite{yang2019}. Yang et al. proposed PresRecRF \cite{yang2024}, which
integrates semantic knowledge of TCM, while Dong et al. presented PresRecST \cite{dong2024}, which combines syndrome differentiation and treatment planning. These studies, mainly supported by open source databases such as TCMSP \cite{ru2014tcmsp} and TCMID \cite{xue2012tcmid}, attempt to elucidate the intricate therapeutic mechanisms of herbal medicines at the microscopic level, advancing the standardization and modernization of TCM.

Despite these advances, current methods for recommending TCM formulas still have some limitations. Many previous studies often overlook the inherent properties of herbs that are crucial in TCM formulas, such as the chemical composition and molecular-scale feature of herbs, only focusing on the textual associations between symptoms and herbs based on a dataset. Some studies have attempted to exploit the deep level relations of symptoms and herbs, but their methods simply integrate the features in different levels separately, thus failing to accurately and fully utilize the comprehensive associations between symptoms and herbs. More importantly, most existing models view the generation of TCM formulas as an unordered prediction task, which can lead to the misuse of herbs with opposite therapeutic properties during the formula generation process, resulting in significant limitations in clinical practice. For example, when a set of identical symptoms in the training data belong to two mutually exclusive TCM syndromes (e.g., cough and runny nose belong to both the type of wind-cold syndrome and the type of wind-heat syndrome) and correspond to different prescriptions, these models could generate a jumbled combination of herbs derived from two diametrically opposed therapeutic methods (such as clearing heat and dispersing cold), thus violating the coherent syndrome differentiation principle of TCM and potentially compromising clinical efficacy.

\begin{figure*}[ht]
  \centering
  \includegraphics[width=\textwidth]{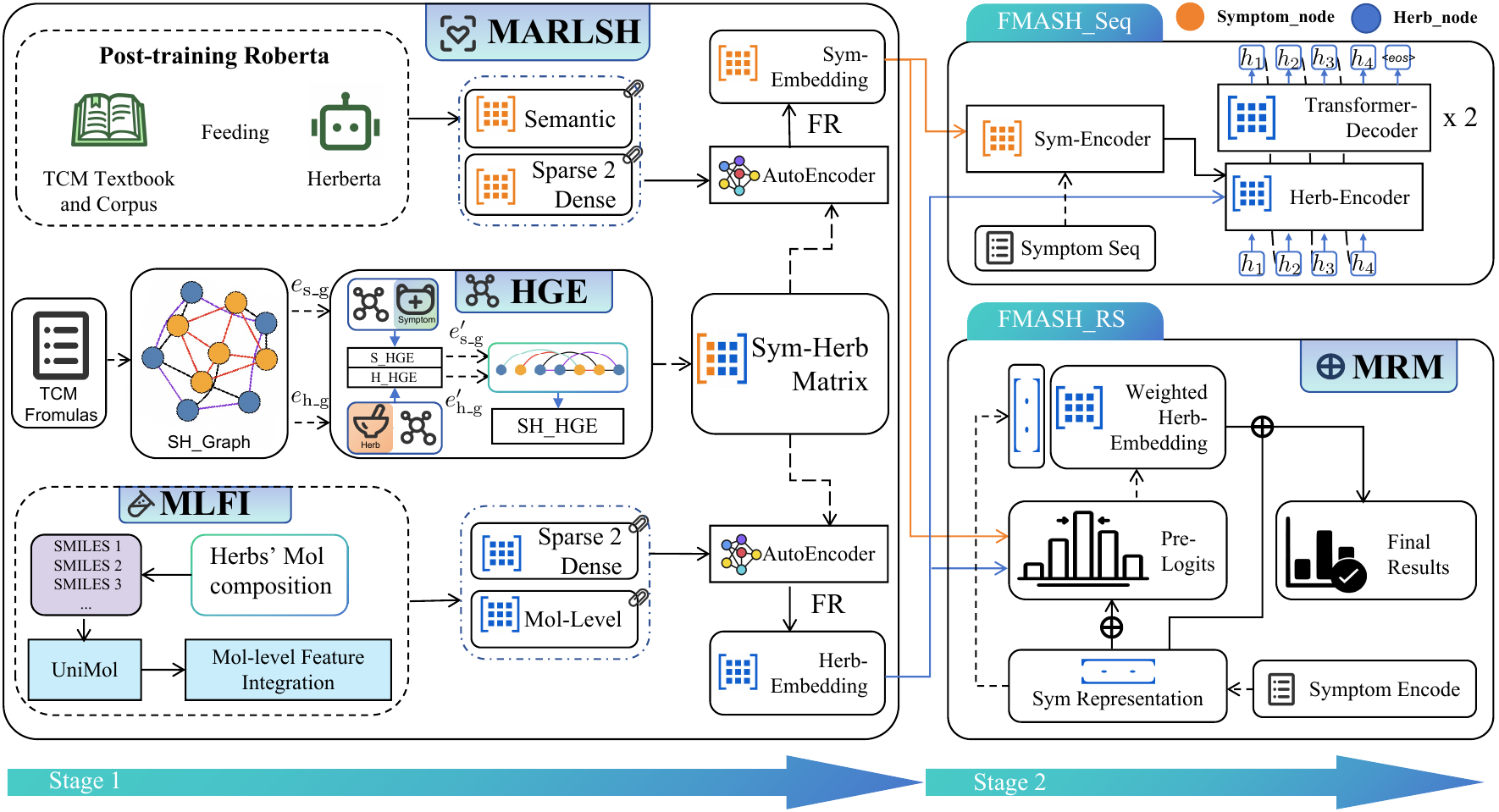}  
\caption{The framework of \textbf{FMASH} is composed of two stages, designated as Stages 1 and 2. In Stage 1, symptom-herb multiscale associations and features are integrated and embedded. In Stage 2, two distinct TCM formula generation models, \textbf{FMASH\_RS} and \textbf{FMASH\_Seq} are trained, respectively. The \textbf{HGE} provides unified graph embeddings for macroscopic hierarchical relations of symptoms and herbs. The \textbf{MLFI} is designed to characterize the functions and features of herbs at the molecular scale. The \textbf{MRM} enhances the integration and utilization of the global features of symptoms and herbs. Feature Refinement (\textbf{FR}) is a module that improves the multiscale features embeddings of symptoms and herbs.}  
\label{fig:fmash_framework}
\end{figure*}

\subsection{Overview of FMASH}
To address these pressing issues, we propose FMASH, a novel framework consists mainly of two stages (designated as Stages 1 and 2) and several functional modules, as shown in Figure \ref{fig:fmash_framework}. In the framework, Stage 1 is mainly for the development of a universal mechanism for \textbf{Multiscale Association Representation Learning of Symptoms and Herbs (MARLSH)}, which can efficiently integrate multiscale associations of symptoms and herbs to provide refined embeddings of them. Stage 2 involves designing and training two different types of model for TCM formula recommendation, i.e., the traditional unordered formula recommendation model (\textbf{FMASH$\_$RS}) and the herb sequential generative model (\textbf{FMASH$\_$Seq}), respectively, based on the refined feature embeddings extracted from Stage 1. In addition, this framework comprises four main functional modules: (1) \textbf{Heterogeneous Graph Embedding (HGE)} module, (2) \textbf{Molecular-Level Feature Integration (MLFI)} module, (3) \textbf{Matching Refinement Mechanism (MRM)} module, and (4) \textbf{Feature Refinement (FR)} module. The main contributions of this work can be summarized as follows:

\begin{itemize}

\item \textbf{MARLSH} is developed to integrate and exploit complicated multiscale associations and features of symptoms and herbs from original datasets of the TCM corpus, formulas, and molecular compositions, providing effective embeddings of multiscale features in a unified semantic space. This mechanism, in conjunction with \textbf{FR} that reduces computational complexity and improves the model's feature embedding and generalization ability, is compatible with both the traditional unordered "many-to-many" herb recommendation task and the serialized herb generation task through the unified representation layer, significantly enhancing its usability.
    
\item The proposed \textbf{HGE} effectively captures the hierarchical local and global relations in heterogeneous graph networks of symptoms and herbs, providing the effective unified embeddings of their multi-type features. This method first aggregates local relations via GCNs and then propagates long-range relations via Mamba\cite{gu2024mamba}, allowing the method to achieve a more nuanced and powerful utilization of the complex relations within the graphs.
    
\item The developed \textbf{MLFI} effectively characterizes herbal medicines at the micro-molecular level with the integration of their macroscopic properties (including the four natures, five flavors, etc.), improving the interpretable and refined embedding of herbs. This method utilizes attention mechanisms to obtain molecular-level representation of herbs through supervised learning, and utilizes the proposed \textbf{Latent Mol} method to address the issues of missing data in the molecular composition of herbs.   

\item Based on the refined embeddings obtained from MARLSH, \textbf{FMASH$\_$RS} and \textbf{FMASH$\_$Seq} are designed and trained, respectively. Both models demonstrate superior performance on the TCM formula recommendation task compared to the current SOTA model. Meanwhile, \textbf{MRM} has been developed to improve the utilization of multiscale associations of symptoms and herbs, promoting the performance of \textbf{FMASH$\_$RS}. 
 
\end{itemize}

The developed method demonstrates advantages in the generation and recommendation of TCM formulas (output) based on given symptoms (input). By effectively integrating multiscale associations of symptoms and herbs, it  enhances the application of the AI-based TCM formula recommendation system. The experimental results demonstrate that the FMASH framework-based model outperforms the state-of-the-art (SOTA) model on both datasets. Our model shows improvements of 3.38\% in Precision@5, 3.89\% in Recall@5 and 3.69\% in F1@5 in Dataset1. In Dataset2, Precision@5, Recall@5 and F1@5 increase by 2.64\%, 1.92\%, and 2.23\%, respectively, compared to the SOTA model. The results confirm the effective capability of FMASH in the TCM formula recommendation task. 

The structure of this paper is organized as follows: Section \ref{sec:related work} discusses related work on AI-based TCM recommendation systems. Section \ref{sec:method} describes the methodology and construction of FMASH in detail. Section \ref{sec:experiment} presents the experimental results and the corresponding analysis. Finally, Section \ref{sec:summary} summarizes the paper and discusses
the direction of future research.

\section{Related Work}
\label{sec:related work}
\subsection{Topic Modeling-Based Approaches}
Topic modeling has long been used in herbal recommendation and remains a fundamental technique to uncover latent patterns in TCM prescriptions. The PTM model reveals the logical relationships between symptoms and herbs \cite{yao2018}, while KGETM enhances the representation of the interaction  between symptoms and herbs through knowledge graph embeddings \cite{wang2019}. However, these models often rely on the bag-of-words framework, which struggles to capture the complex semantic relationships inherent in TCM short texts. Although recent improvements have incorporated domain-specific knowledge, such as herbal compatibility rules, challenges persist in terms of computational complexity and scalability \cite{wang2019}.

\subsection{Sequence Generation-Based Approaches}
Deep learning advancements have paved the way for the application of sequence generation models in herbal recommendation. For example, the TCM Translator treats herbal recommendation as a machine translation task, using Transformer architectures to map symptoms to prescription sequences \cite{wang2019tcm}. Similarly, AttentiveHerb \cite{liu2019} uses dual-attention mechanisms to distinguish the roles of primary and secondary symptoms in recommendations \cite{wenbo2025lamgcn}. Despite these developments, many models still neglect the chemical properties of herbs, limiting their precision and interpretability \cite{yang2024}.

\subsection{Graph Neural Network-Based Approaches}

Graph neural networks (GNNs) have become a dominant paradigm for the recommendation of TCM formulas due to their ability to model the relational structure of the symptom-herb interactions \cite{velickovic2017graph, li2022heterogeneous}. Early works in this line include SMGCN \cite{jin2020}, which constructed heterogeneous graphs of symptom-herb co-occurrence and employed graph convolutional  layers to propagate neighborhood information, and KDHR \cite{yang2019}, which enriched the graph with knowledge-driven attribute embeddings covering efficacy, toxicity, and meridian tropism.

Building on these foundations, various models have sought to strengthen the expressiveness of graph-based representations from complementary angles. SMRGAT \cite{yang2023smrgat} employed the residual graph layers to process the relations between symptoms and herbs and used a multi-head attention mechanism to differentiate their effect. This model integrated herbal attributes to derive 23-dimensional property vectors of the herb and obtain the final herb embeddings. SCEIKG \cite{liu2024sceikg} considered the patient’s status and symptoms over time and extended the GNN paradigm along the temporal development of the patient's status, taking the herb recommendation task as a sequential decision process during multiple clinical visits. PresRecST \cite{dong2024} combined a TCM knowledge graph with residual neural networks to implement the workflow for the syndrome differentiation and treatment planning. TCMRGCL \cite{hu2025tcmrgcl} overcame the bottleneck of data sparsity and the long-tail distribution of node popularity, through contrastive pre-training. This model constructed two augmented views of the symptom-herb graph for contrastive pre-training in order to obtain initial symptom and herb embeddings.

Despite these considerable improvements, all of the above GNN-based models still only work at a symbolic level. They represent herbs through co-occurrence statistics, categorical property labels, or knowledge-graph triples, without leveraging the three-dimensional molecular structures that ultimately govern pharmacological activity. Consequently, the chemical basis for therapeutic interactions between  symptoms and herbs remains to be explored to a sufficient extent \cite{yang2024}.

\subsection{Molecular Representation Learning Approaches}
UniMol is a universal framework for molecular representation learning that efficiently captures three-dimensional spatial information of molecules \cite{zhou2023unimol}. Although it was not originally designed for studies related to Traditional Chinese Medicine (TCM), its application in modern drug discovery demonstrates its potential for herbal recommendation research. Models inspired by Uni-Mol, by integrating molecular data with symptom-semantic associations, can significantly improve the precision, robustness, and interpretability of herbal recommendation systems \cite{liu2021gem, wang2022gem2}.

\section{Methods}
\label{sec:method}
In this section, the construction and implementation of each main functional module of the FMASH framework are systematically described.

\subsection{Heterogeneous Graph Embedding}
\begin{figure}[ht] 
  \centering
  \includegraphics[width=0.9\columnwidth]{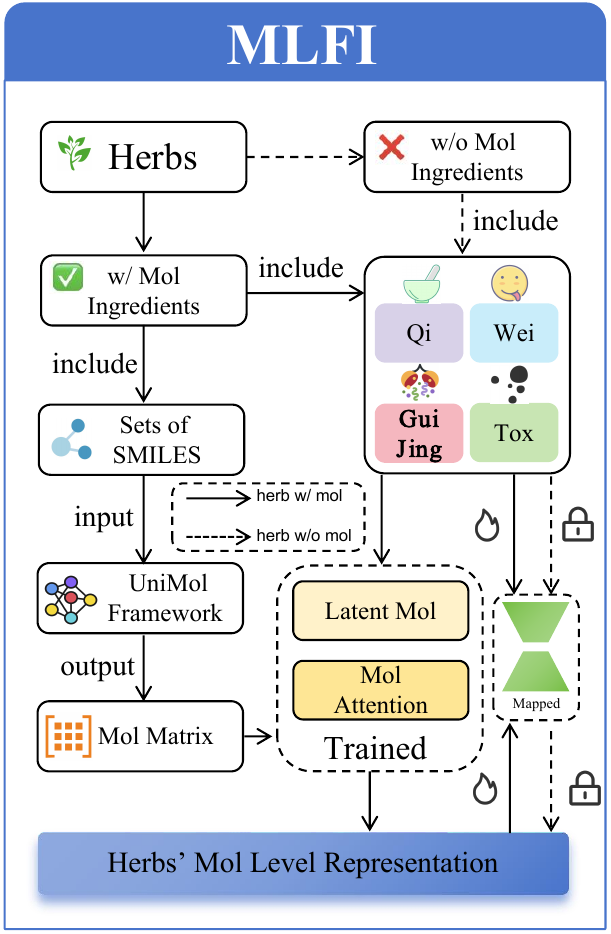} 
  \caption{The Framework of Molecular-Level Feature Integration method, which enhances the refined representation of multiscale herb features through effectively integrating molecular-scale chemical characteristics and macroscopic properties of herbs. }
  \label{fig: MLFI_framework}
\end{figure}

To effectively model the hierarchical multi-type relations within the heterogeneous graph of symptoms and herbs, we introduce a novel \textbf{Heterogeneous Graph Embedding (HGE)} method. This approach structures the learning process to first aggregate local neighborhood information and then capture long-range relations, all within a unified paradigm.

Let the overall heterogeneous graph be denoted as \(G = (V, E)\), where \(V = V_{\text{sym}} \cup V_{\text{herb}}\) is the set of symptom and herb nodes, and \(\mathbf{X} \in \mathbb{R}^{|V| \times d}\) is the characteristic matrix of the initial nodes. Initially, the method isolates homogeneous subgraphs, such as the herb-herb subgraph \(G_{\text{herb-herb}} = (V_{\text{herb}}, E_{\text{herb-herb}})\) and the symptom-symptom subgraph \(G_{\text{sym-sym}} = (V_{\text{sym}}, E_{\text{sym-sym}})\). For each subgraph, a Graph Convolutional Network (GCN) layer is applied to capture local neighborhood features among nodes of the same type. For example, for the herb subgraph with features \(\mathbf{X}_{\text{herb}}\), the GCN layer produces locally-aware features \(\mathbf{X}_{\text{gcn}, \text{herb}}\). Thus, a structurally-ordered sequence is created by sorting these nodes based on their degrees \(\mathbf{d}_{\text{herb-herb}}\) via a permutation \(\mathbf{p}_{\text{herb-herb}}\). This sequence is then processed by a bidirectional Mamba model\cite{wang2024graph}, \(\text{Mamba}_{\text{herb-herb}}\), to model long-range relations and produce enhanced features \(\mathbf{X}'_{\text{herb}}\). The same processing procedure is applied in parallel to the subgraphs of the symptoms. This enhancement process for a generic subgraph can be represented as follows:
\begin{equation}
\setlength\abovedisplayskip{8pt}
\mathbf{X}'_s = \text{Unsort}_{\mathbf{p}_s}(\text{Mamba}_s(\text{Sort}_{\mathbf{p}_s}(GCN_s(\mathbf{X}_s, \mathbf{E}_s)))).
\setlength\belowdisplayskip{8pt}
\end{equation}

According to this, the features extracted from both subgraphs are concatenated to form a new, context-rich feature matrix for the entire graph, \(\mathbf{X}_e = \text{concat}(\mathbf{X}'_{\text{herb}}, \mathbf{X}'_{\text{sym}})\). This representation embeddings are first passed through a GCN layer to refine features based on the overall graph structure. Then, they are converted into a sequence by sorting all nodes according to their degrees \(\mathbf{d}_g\) and processed with a bidirectional Mamba model \(\text{Mamba}_g\). Thus, the final output, \(\mathbf{X}'' \in \mathbb{R}^{|V| \times d}\), represents the final embeddings of nodes that have hierarchically integrated local and global relations of symptoms and herbs. The transformation process is given by:
\begin{equation}
\setlength\abovedisplayskip{8pt}
\mathbf{X}'' = \text{Unsort}_{\mathbf{p}_g}(\text{Mamba}_g(\text{Sort}_{\mathbf{p}_g}(\text{GCN}_g(\mathbf{X}_e, \mathbf{E})))).
\setlength\belowdisplayskip{8pt}
\end{equation}

This method applies GCNs separately on symptom-symptom and herb-herb subgraphs to capture local neighborhood features among nodes of the same type, and then utilizes bidirectional Mamba layers to propagate long-range relations through the aggregated node sequences. Finally, the two groups of features are merged back into a heterogeneous graph embedding, enabling further modeling and representation of local and global features.

\subsection{Molecular-Level Feature Integration}

In order to effectively integrate and utilize the molecular-scale chemical characteristics and macroscopic properties of herbs, as well as overcome the significant challenge of missing molecular information of herbs in a dataset, the Molecular-Level Feature Integration (MLFI) method is developed. This method, as shown in Figure \ref{fig: MLFI_framework}, provides a unified framework to generate a molecular representation relevant to a single function \(\mathbf{v}_{h, \text{final}}\) that can adaptively evolve one of the following two processes for each herb, according to the availability of data. 

\textbf{One process is for herbs with partially missing molecular composition data}. To address this problem, we have proposed the \textbf{Latent Mol} method. This method first aggregates the known molecular embeddings via a property-guided attention mechanism. Subsequently, a gating network is employed to dynamically fuse this bottom-up representation to capture latent or unrecorded ingredients. The process is initiated by encoding each known molecule from its SMILES string into an embedding vector \(\mathbf{e}_{h,k}\) using the UniMol\cite{zhou2023unimol} framework. Subsequently, these elements are aggregated through property-guided attention to form an initial representation of the herb \(\mathbf{v}_h\):
\begin{equation}
\begin{split}
    \mathbf{v}_h &= \sum_{k \in \mathcal{M}_h} \alpha_{h,k} \mathbf{e}_{h,k}, \\
    \alpha_{h,k} &= \mathrm{softmax}_k \left( \frac{(\mathbf{W}_q \mathbf{p}_h)^\top (\mathbf{W}_k \mathbf{e}_{h,k})}{\sqrt{d_k}} \right).
\end{split}
\end{equation}

Next, a gating network fuses this vector with a holistic and easily learnable embedding \(\mathbf{h}_e \in \mathbb{R}^{d}\) for the herb. Unlike \(\mathbf{v}_h\) which is calculated from known molecules, \(\mathbf{h}_e\) is a trainable vector that is randomly initialized and optimized during training through gradient descent, allowing it to capture latent or unrecorded ingredient information. Here, the gate value \(\lambda_h \in [0, 1]\) allows the model to rely more on \(\mathbf{h}_e\), if the known molecular information of the herb is weak. The final representation vector for the herb is calculated as:
\begin{equation}
    \lambda_h = \sigma(\mathbf{W}_g \mathbf{v}_h + \mathbf{b}_g)
\end{equation}
\begin{equation}
    \mathbf{v}_{h, \text{final}} = \lambda_h \odot \mathbf{v}_h + (1 - \lambda_h) \odot \mathbf{h}_e
\end{equation}
where \(\sigma\) is the sigmoid function and \(\mathbf{W}_g, \mathbf{b}_g\) are parameters of the gating network. This unified approach ensures that every herb is assigned a final embedding vector that is robust to the potential incompleteness of its molecular data.

\textbf{Another process is for herbs with entirely missing molecular composition data}. Inspired by the idea of Variational Autoencoder (VAE) \cite{kingma2013auto}, we have trained a VAE-like approach to address this issue and generate these missing representations. This approach learns a probabilistic mapping from an herb's classical 23-dimensional property vector \cite{yang2023smrgat} \(\mathbf{p}_h\) to the latent distribution of its molecular features, using the subset of herbs with complete data for training. The VAE-like approach is trained to optimize the standard Evidence Lower Bound Objective (ELBO) by convention:
\begin{equation}
\mathcal{L}_{\text{VAE}} = \mathbb{E}_{q_{\phi}(\mathbf{z} | \mathbf{p}_h)}[\log p_{\theta}(\mathbf{v}_h | \mathbf{z})] - D_{\text{KL}}(q_{\phi}(\mathbf{z} | \mathbf{p}_h) || p(\mathbf{z}))
\label{eq:VAE}
\end{equation}
In this formulation, the encoder \(q_{\phi}\) maps an input property vector \(\mathbf{p}_h\) to a latent variable \(\mathbf{z}\), and the decoder \(p_{\theta}\) reconstructs the aggregated molecular representation \(\mathbf{v}_h\) from \(\mathbf{z}\). The training and imputation processes are detailed in Algorithm \ref{alg:vae_imputation}. After training, the decoder can generate a molecular embedding \(\mathbf{v}_h\) for any herb given only its property vector, effectively complementing missing molecular-level data from the herb and also enhancing the integration of features. The performance of the approach will be demonstrated in the experiment later.

\begin{algorithm}[htbp]
\caption{\textbf{Molecular-level features integration and VAE training}}
\label{alg:vae_imputation}
\begin{algorithmic}[1]
\STATE \textbf{Phase1: VAE Training}
\STATE \quad \textbf{Train} a VAE model \((\text{Encoder } q_{\phi}, \text{Decoder } p_{\theta})\) on a dataset of complete pairs \(D_{\text{complete}} = \{(\mathbf{p}_h, \mathbf{v}_h)\}\) by minimizing the loss \(\mathcal{L}_{\text{VAE}}\) from Equation \ref{eq:VAE}.
\STATE
\STATE \textbf{Phase2: Feature enhancement via Trained VAE}
\STATE \quad \textbf{Input:} A property vector \(\mathbf{p}_{h, \text{missing}}\) for a herb with missing data; the trained VAE \((q_{\phi}, p_{\theta})\).
\STATE \quad \textbf{Output:} The imputed molecular embedding \(\mathbf{v}_{h, \text{imputed}}\).
\STATE \quad \((\mu, \sigma) \leftarrow q_{\phi}(\mathbf{p}_{h, \text{missing}})\) \COMMENT{Encode properties to latent space}
\STATE \quad \(\mathbf{z} \sim \mathcal{N}(\mu, \sigma^2)\) \COMMENT{Sample a latent vector}
\STATE \quad \(\mathbf{v}_{h, \text{complete}} \leftarrow p_{\theta}(\mathbf{z})\) \COMMENT{Decode to generate feature}
\STATE \quad \textbf{Return} \(\mathbf{v}_{h, \text{completed}}\)
\end{algorithmic}
\end{algorithm}

\subsection{Feature Refinement}
During the training process, the features of the symptoms and herbs are compressed into embedding vectors with a given dimension. A good compression embedding method directly affects the model's performance and the related computational complexity. To reduce computational complexity and improve the ability of feature embedding and generalization of the model, the Feature Refinement (FR) approach based on AutoEncoder\cite{hinton2006reducing} has been developed for the precise evaluation of loss of feature compression and reconstruction. Specifically, the FR is estimated by comparing and calculating the Mean Squared Error (MSE) between the compression and reconstruction output with the original input. According to FR, the model can be trained to accurately compress the high-dimensional representations of symptoms and herbs into 64-dimensional feature embeddings that comprehensively integrate the features of symptoms and herbs.

\subsection{Design of FMASH\_RS}
After obtaining the embedding of the multiscale features of the symptoms and herbs using MARLSH, we develop a recommendation model (shown in the lower part of Stage 2 in Figure \ref{fig:fmash_framework}) to derive the matching coefficients between the herbs and the symptoms. Then, based on the input of symptoms, FMASH\_RS sorts the herbs in descending order of the matching coefficients and recommends the appropriate herbs.

Traditionally, the matching coefficients are calculated through the inner product between the symptom embedding and the herb embedding matrix. However, the traditional way of calculating the matching coefficients could not be sufficient to capture the complex, non-linear, and long-range relations between symptoms and herbs. To address this, we develop the \textbf{Matching Refinement Mechanism (MRM)} to improve the calculation of the matching coefficients between symptoms and herbs. It can utilize complex long-range relations between symptoms and herbs, enhancing the integration of their various features. 

First, the predicted scores of the candidate herbs obtained from the previous stage are used to weight the corresponding herb features, which serves as a soft re-ranking strategy for herb representations. Let $\mathbf{H} \in \mathbb{R}^{d \times N}$ denote the embedding matrix of all candidate herbs, where $d$ is the embedding dimension and $N$ is the number of candidate herbs. Let $\mathbf{p}=[p_1,p_2,\ldots,p_N]^\top \in \mathbb{R}^{N}$ denote the corresponding weight vector, where $p_i$ represents the predicted probability or normalized score of the $i$-th herb. Each herb token is scaled column-wise by its corresponding weight, resulting in the weighted embedding matrix of herb features:
\begin{equation}
\bar{\mathbf{H}}=\mathbf{H}\,\mathrm{diag}(\mathbf{p}) \in \mathbb{R}^{d \times N},
\end{equation}
where $\mathrm{diag}(\mathbf{p}) \in \mathbb{R}^{N \times N}$ denotes the diagonal matrix constructed from $\mathbf{p}$. Thus, the $i$-th column of the matrix $\bar{\mathbf{H}}=$ represents the weighted feature embedding of the $i$-th herb token, which is
\begin{equation}
\bar{\mathbf{h}}_i = p_i \mathbf{h}_i,\quad i=1,2,\ldots,N.
\end{equation}
Unlike approaches that compress all herb features into a single vector, this design retains all herb tokens while enhancing the representation of herbs with higher weights and suppressing interference from those with lower weights, thereby providing richer information for subsequent fine-grained modeling.

To further model the global associations between symptoms and all candidate herbs, a Transformer Encoder is used to fuse the symptom token with the sequence of weighted herb tokens. Let $\mathbf{s} \in \mathbb{R}^{d}$ denote the aggregated symptom token, and let $\mathbf{c}_{\mathrm{cls}} \in \mathbb{R}^{d}$ denote a learnable classification (CLS) token. Then, all the tokens are concatenated in sequence to form the input of the Transformer with the CLS token placed at the end, which can be written as:
\begin{equation}
\mathbf{X}=[\mathbf{s},\bar{\mathbf{H}},\mathbf{c}_{\mathrm{cls}}] \in \mathbb{R}^{d \times (N+2)}.
\end{equation}
The resulting sequence of tokens is then fed into the Transformer Encoder for the purpose of contextual representation learning:
\begin{equation}
\mathbf{Z}=\mathrm{TransEnc}(\mathbf{X}) \in \mathbb{R}^{d \times (N+2)}.
\end{equation}
Finally, the hidden state corresponding to the CLS token in the last position is extracted as the global joint representation of symptoms and herbs, denoted by $\mathbf{z}_{\mathrm{cls}} \in \mathbb{R}^{d}$. Based on this representation, the final prediction Scores are obtained through a linear projection:
\begin{equation}
\mathrm{Scores}=\mathbf{W}\mathbf{z}_{\mathrm{cls}}+\mathbf{b},
\end{equation}
where $\mathbf{W}$ and $\mathbf{b}$ are learnable parameters. This mechanism not only performs soft selection on candidate herbs through prediction-aware weighting, but also precisely guides the model to focus on more important features and relations between symptoms and herbs, improving the model's performance on herb recommendation.

\subsection{Design of FMASH\_Seq}
Traditional recommendation systems may erroneously generate a recommended list by combining elements from two different ground truths. When applied in the field of herbal medicine, this phenomenon can result in a combination of mutually exclusive herbs. For example, two distinct TCM formulas might be individually effective, but the mixture of herbs from both could not only compromise the therapeutic effect but also potentially introduce toxic side effects. The sequence-to-sequence paradigm mitigates this risk by imposing a strict sequential dependency among the generated herbs, thus preserving the coherent syndrome differentiation principle of TCM.

Therefore, based on the refined multiscale embeddings produced by MARLSH, we develop \textbf{FMASH\_Seq} (shown in the upper part of Stage 2 in Figure \ref{fig:fmash_framework}), a sequence-to-sequence model that auto-regressively generates an ordered herbal formula. Unlike conventional Transformer-based baselines that operate on single-scale embeddings, FMASH\_Seq explicitly leverages the refined representations of symptoms and herbs obtained from Stage 1 to establish a semantically rich foundation for sequential generation. The architecture consists of three components: a \emph{Symptom Encoder}, a \emph{Herb Encoder}, and $N$ stacked \emph{Transformer Decoder} layers.

\paragraph{Symptom Encoder}
Let $\mathcal{S}=\{s_1,\dots,s_{|\mathcal{S}|}\}$ denote the set of input symptoms. The Symptom Encoder applies multi-head self-attention over the refined symptom embeddings $\{\mathbf{h}_{s_i}\}$ to capture the inter-dependence among symptoms:
\begin{equation}
    \mathbf{M}_s = \mathrm{MultiHead}(\mathbf{H}_s,\,\mathbf{H}_s,\,\mathbf{H}_s) \;\in\;\mathbb{R}^{|\mathcal{S}|\times d},
    \label{eq:sym_encoder}
\end{equation}
where $\mathbf{H}_s=[\mathbf{h}_{s_1};\dots;\mathbf{h}_{s_{|\mathcal{S}|}}]\in\mathbb{R}^{|\mathcal{S}|\times d}$ denotes the embedded matrix of stacked symptoms. Then, a global context vector of symptoms is then obtained through mean pooling:
\begin{equation}
    \mathbf{c} = \frac{1}{|\mathcal{S}|}\sum_{i=1}^{|\mathcal{S}|} \mathbf{m}_{s_i},
    \label{eq:context}
\end{equation}
where $\mathbf{m}_{s_i}$ is the $i$-th row of $\mathbf{M}_s$.

\paragraph{Herb Encoder}
The Herb Encoder serves as a contextual integration module that fuses initial herb features with symptom-level information at each decoding step. Specifically, at step $t$, the previously generated herb refined embedding $\mathbf{h}_{h_{t-1}}$ is first projected and combined with the context vector of symptoms $\mathbf{c}$ through a gated fusion mechanism:
\begin{equation}
    \mathbf{g}_t = \sigma\bigl(\mathbf{W}_g[\mathbf{h}_{h_{t-1}};\,\mathbf{c}] + \mathbf{b}_g\bigr),
    \label{eq:herb_gate}
\end{equation}
\begin{equation}
    \mathbf{d}_t = \mathbf{g}_t \odot \mathbf{W}_h\mathbf{h}_{h_{t-1}} + (1-\mathbf{g}_t) \odot \mathbf{W}_c\mathbf{c},
    \label{eq:herb_decoder}
\end{equation}
where $\sigma(\cdot)$ denotes the sigmoid function, $\mathbf{W}_g \in \mathbb{R}^{d \times 2d}$, $\mathbf{W}_h, \mathbf{W}_c \in \mathbb{R}^{d \times d}$ are learnable parameters, and $\odot$ denotes element-wise multiplication. The gate function $\mathbf{g}_t$ dynamically balances the contribution of the herb-specific information and the symptom context, allowing the model to adaptively weight the importance of each source at different generation steps. This explicit herb-symptom fusion step is absent in the Transformer-based baseline, where herb embeddings are fed directly into the decoder without contextual conditioning. For the first step ($t\!=\!1$), a learnable start-of-sequence token $\mathbf{h}_{\langle\mathrm{sos}\rangle}$ is used.

\paragraph{Transformer Decoder ($\times N$)}
The output of the Herb Encoder is subsequently fed into $N$ stacked Transformer Decoder layers ($N=6$ in our implementation). Each layer $\ell$ applies causal-masked self-attention on the herb representations generated previously, followed by cross-attention on the representation of symptom sequence $\mathbf{M}_s$:
\begin{align}
    \widetilde{\mathbf{D}}_t^{(\ell)} &= \mathrm{MaskedMultiHead}\bigl(\mathbf{D}_{\le t}^{(\ell-1)},\,\mathbf{D}_{\le t}^{(\ell-1)},\,\mathbf{D}_{\le t}^{(\ell-1)}\bigr), \label{eq:causal_self_attn}\\[4pt]
    \mathbf{D}_t^{(\ell)} &= \mathrm{MultiHead}\bigl(\widetilde{\mathbf{D}}_t^{(\ell)},\,\mathbf{M}_s,\,\mathbf{M}_s\bigr), \label{eq:cross_attn}
\end{align}
where $\mathbf{D}_{\le t}^{(0)}=[\mathbf{d}_1;\dots;\mathbf{d}_t]$ collects the Herb Encoder output up to step $t$. Notably, the cross-attention mechanism in Eq.~\eqref{eq:cross_attn} allows each decoding step to dynamically attend to different symptoms in $\mathbf{M}_s$, enabling the model to adaptively focus on the most relevant clinical information at each stage of formula generation. Since $\mathbf{M}_s$ carries the multiscale associations refined by MARLSH, this cross-attention captures richer symptom-herb interactions compared to the Transformer-based baseline.

\paragraph{Output distribution}
After the final Transformer Decoder layer, a scoring function computes the logit for each candidate herb $h$ at step $t$:
\begin{equation}
    \phi_t(h) = \mathbf{W}_o\bigl[\mathbf{d}_t^{(N)};\,\mathbf{h}_h\bigr] + \mathbf{b}_o,
    \label{eq:score}
\end{equation}
where $\mathbf{d}_t^{(N)} \in \mathbb{R}^{d}$ is the output of the $N$-th Transformer Decoder layer, $\mathbf{h}_h \in \mathbb{R}^{d}$ is the refined herb embedding, $\mathbf{W}_o \in \mathbb{R}^{1 \times 2d}$ and $\mathbf{b}_o \in \mathbb{R}$ are learnable parameters. The concatenation $[\mathbf{d}_t^{(N)};\,\mathbf{h}_h] \in \mathbb{R}^{2d}$ enables the model to jointly consider the decoded contextual information and the intrinsic multiscale properties of each candidate herb. A masked softmax is then applied to ensure that previously selected herbs are excluded from the candidate set:
\begin{equation}
    P(h_t=h\mid h_{<t},\,\mathcal{S})
    = \frac{\exp(\phi_t(h))\cdot\mathbf{1}[h\notin\{h_1,\dots,h_{t-1}\}]}{\sum_{h'}\exp(\phi_t(h'))\cdot\mathbf{1}[h'\notin\{h_1,\dots,h_{t-1}\}]}.
    \label{eq:masked_softmax}
\end{equation}
This prevents redundant recommendations and maintains the compositional coherence of the generated formula.

\paragraph{Training}
Given a ground-truth formula $\mathbf{h}^{*}=(h_1^{*},\dots,h_T^{*})$ with the number of $T$ herbs, the sequential model is trained with teacher forcing under the standard cross-entropy loss:
\begin{equation}
    \mathcal{L}_{\mathrm{seq}}
    = -\frac{1}{T}\sum_{t=1}^{T}\log P(h_t^{*}\mid h_{<t}^{*},\,\mathcal{S}).
    \label{eq:loss_seq}
\end{equation}

\paragraph{Inference}
During inference, herbs are generated auto-regressively through greedy decoding: $h_t=\arg\max_{h}P(h_t=h\mid h_{<t},\mathcal{S})$. A special end-of-sequence token $\langle\mathrm{eos}\rangle$ is appended to each ground-truth formula during training, allowing the model to learn when to terminate generation. At test time, the decoding stops when $\langle\mathrm{eos}\rangle$ is selected or a maximum length is reached. This mechanism enables FMASH\_Seq to produce variable-length formulas that respect the compositional coherence required by the TCM practice.

\section{Experiments}
\label{sec:experiment}
In this section, we detail the experimental setup and dataset. The baseline models and evaluation metrics are introduced. The experiments on model's performance, the ablation study, and the case study are conducted. Finally, the experimental results are presented and analyzed.

\subsection{Experimental Settings}

\paragraph{Configuration}
All experiments are conducted on an Intel Platinum 8352V CPU and an NVIDIA RTX A6000 GPU, using Python~3.10.0 and PyTorch~1.13.1+cu121.

\paragraph{Hyperparameters}
The model is trained for 200 epochs with a batch size of 32 and a dropout rate of 0.1. Adam optimizer was utilized, with a learning rate of $1\times10^{-4}$, a weight decay of $7\times10^{-3}$, and a StepLR scheduler (step size 7, $\gamma=0.8$). The random seed has been set to 2025 to maintain the chronological sequence. The embedding dimension is set to $d=64$; the heterogeneous graph contains 390 symptom nodes and 805 herb nodes, for a total of 1,195 nodes. For hierarchical reasoning, the length of the embeddings for the syndrome and treatment is set at 50 and 75, respectively. For multimodal alignment, two autoencoders are used to compress the embedding vectors: the symptom autoencoder $1152\!\rightarrow\!64$ and the herb autoencoder $640\!\rightarrow\!64$.

\subsection{Datasets and Baseline Models}
\paragraph{Datasets}
We use two commonly used public datasets in our experiments to exhibit the performance of our proposed model. The first dataset, referred to as \textbf{Dataset1}, was original from KDHR \cite{yang2019} and has since been adopted as a standard benchmark in several studies. The second dataset, referred to as \textbf{Dataset2}, was adopted from SMGCN \cite{jin2020}, which is compiled from classical texts of TCM and clinical data. Dataset1 contains 33,765 prescription records that cover 390 distinct symptoms and 805 herbs. Following the same partition protocol used by Hu et al. \cite{hu2025tcmrgcl}, we randomly divide Dataset1 into a training set of 20,259 prescriptions and a test set of 13,506 prescriptions in the experiment.
For Dataset2, which comprises 26,360 prescription records that cover 360 symptoms and 753 herbs, we directly adopt the original train/test partition released by Jin et al. \cite{jin2020}, producing 22,917 training prescriptions and 3,443 test prescriptions. For both datasets, we obtain molecular-level features of herbs through our developed MLFI module and encode the semantics of symptoms using RoBERTa \cite{liu2019roberta}. 

\paragraph{Baselines}
We compare FMASH\_RS and FMASH\_Seq with the following baseline models: SMGCN, KDHR, SMRGAT, SCEIKG, PresRecST, and TCMRGCL. The introduction about these models can be found in Section \ref{sec:related work}, and the literature \cite{hu2025tcmrgcl} can also be referenced for more information. To ensure a fair comparison, the performance of baseline models on two datasets is directly inherited from the literature \cite{hu2025tcmrgcl}, while maintaining consistent settings and the same data partitioning in our experiments.

\subsection{Evaluation Metrics}
The evaluation metrics of Precision@$k$, Recall@$k$ and F1-score@$k$ ($k\in\{5,10,20\}$) have been adopted in the measurement of the model's performance. Let $R_k(\mathrm{sym})$ denote the top-$k$ herbs recommended for a set of symptoms, and let $T(\mathrm{sym})$ denote the set of ground-truth herbs. These metrics can be calculated as follows:
\begin{align}
    \text{Precision@}k
    &= \frac{\bigl|R_k(\mathrm{sym})\cap T(\mathrm{sym})\bigr|}{k},\\[4pt]
    \text{Recall@}k
    &= \frac{\bigl|R_k(\mathrm{sym})\cap T(\mathrm{sym})\bigr|}{\bigl|T(\mathrm{sym})\bigr|},\\[4pt]
    \text{F1-score@}k
    &= 2\cdot\frac{\text{Precision@}k \cdot \text{Recall@}k}{\text{Precision@}k + \text{Recall@}k}.
\end{align}

However, these conventional recommendation evaluation metrics are not suitable for assessing the performance of the sequential generative task of the TCM formula, due to the sequence-sensitive and combination-critical nature of such tasks. Meanwhile, these traditional metrics cannot reduce the bias caused by the different lengths of the ground-truth herbal formulas. 

Therefore, we adopt normalized \textbf{Matched Precision (MP)} to better evaluate the average performance of sequential generative models for TCM prescription generation. For a given set of symptoms $\mathrm{sym}$, let $R_k(\mathrm{sym})$ denote the set of top-$k$ predicted herbs, and let $\mathcal{T}(\mathrm{sym})=\{T_1(\mathrm{sym}),\dots,T_m(\mathrm{sym})\}$ denote all sets of ground-truth herbs associated with the same input of symptoms (i.e., multiple valid herbal formulas may correspond to the same symptoms in the dataset). MP measures the best precision of the top-$k$ predicted herbs against all ground-truth herbal formulas associated with the same symptoms and is defined as
\begin{equation}
\mathrm{MP@k} \;=\; \max_{1\le i\le m}\frac{\bigl|R_k(\mathrm{sym})\cap T_i(\mathrm{sym})\bigr|}{k}\ .
\end{equation}

In addition, normalized \textbf{Matched Recall (MR)} is adopted to measure how well the top-$k$ predicted herbs cover the ground-truth herbal formulas associated with the same symptoms, while reducing the bias caused by varying formula lengths. Specifically, the denominator is normalized by $\min\{k,|T_i(\mathrm{sym})|\}$, since at most $k$ herbs can be matched by the top-$k$ predictions. MR is defined as
\begin{equation}
\mathrm{MR@k} \;=\; \max_{1\le i\le m}\frac{\bigl|R_k(\mathrm{sym})\cap T_i(\mathrm{sym})\bigr|}{\min\{k,|T_i(\mathrm{sym})|\}}\ .
\end{equation}

Analogously, the normalized \textbf{Matched F1 (MF1)} is proposed based on MP and MR metrics, which can be calculated as: 
\begin{equation}
\mathrm{MF1@}k \;=\; \frac{2\;\cdot\; \mathrm{MP@}k\;\cdot\;\mathrm{MR@}k}{\mathrm{MP@}k \;+\; \mathrm{MR@}k}\ .
\end{equation}


\subsection{Experimental Results}
As shown in Table \ref{tab:rs_result_data1} and Table \ref{tab:rs_result_data2}, the FMASH\_RS model developed based on the proposed FMASH framework demonstrate superior performance compared to other baseline models in both datasets. Specifically, in Dataset1, our FMASH\_RS model has achieved a significant improvement compared to the SOTA model, with increases of 3.38\% in Precision@5, 3.89\% in Recall@5, and 3.69\% in F1-score@5. In Dataset2, our FMASH\_RS model has achieved improvements of 2.64\% in Precision@5, 1.92\% in Recall@5, and 2.23\% in F1-score@5, compared to the SOTA model. This may be because the proposed HGE allows the model to capture captures both local and global hierarchical relationships of symptoms and herbs, and the MLFI effectively integrates the molecular-level feature of herbs, resulting in high-quality and robust features. The combination of these functional modules provides refined representation embeddings of multiscale relations and features of symptoms and herbs, facilitating the performance of the model. 
 
In addition, the performance of the FMASH framework-based herbal sequential generative model, FMASH\_Seq, has been tested on Dataset1. As stated previously, the metrics of Matched Precision (MP), Matched Recall (MR), and Matched F1-score (MF1) are used to circumvent the influence of  various lengths of the ground-truth herbal formulas during model evaluation. The results in Table \ref{tab:seq_result} show that the FMASH\_Seq model achieves significant improvements in all metrics compared to the conventional Transformer-based model. Furthermore, FMASH\_Seq exhibits notable improvements in relevant evaluation metrics for both Top-5 and Top-10 compared to the traditional unordered herbal formula recommendation mode, demonstrating the effectiveness of the FMASH framework. Specifically, it has achieved relative improvements of 28.35\% in MP@5, 22.59\% in MR@5, and 24.96\% in MF1@5 compared to the current SOTA recommendation model, demonstrating its prominent advantages. Due to the inherently sparse and imbalanced nature of herbs and symptoms in the data, which exhibits a long-tailed distribution in their frequency of occurrence \cite{hu2025tcmrgcl}, the FMASH\_Seq model can minimize the generation of incompatible herb combinations to align more closely with the actual principles of TCM treatment. Subsequent case studies further demonstrate the advantages of FMASH\_Seq in addressing such challenges.

It is important to note that, since the two types of model are different in generating and recommending herbal formulas, these experiments are not intended to determine which model is superior, but rather to provide two distinct options in the field of herbal formula recommendation and demonstrate their effectiveness. These experimental results confirm that the two models constructed based on the FMASH framework significantly outperform the existing SOTA models in all evaluation metrics. This fully validates the effectiveness of our proposed framework in addressing the task of automatic recommendations of the TCM formula and lay a solid foundation for further development in this area.

\begin{table*}[!t]
\centering
\small
\setlength{\tabcolsep}{4.5pt}
\renewcommand{\arraystretch}{1.12}
\resizebox{\textwidth}{!}{
\begin{tabular}{|l|ccc|ccc|ccc|}
\hline
\multirow{2}{*}{\textbf{Model}} 
& \multicolumn{3}{c|}{\textbf{Precision}} 
& \multicolumn{3}{c|}{\textbf{Recall}} 
& \multicolumn{3}{c|}{\textbf{F1-score}} \\
\cline{2-10}
& \textbf{P@5}  & \textbf{P@10}  & \textbf{P@20}  
& \textbf{R@5}  & \textbf{R@10}  & \textbf{R@20}
& \textbf{F1@5} & \textbf{F1@10} & \textbf{F1@20} \\
\hline
SMGCN      & 0.1933 & 0.1568  & 0.1153  & 0.1265  & 0.2031  & 0.3027 
& 0.1529  & 0.1770  & 0.1670 \\
KDHR       & 0.2138  & 0.1660  & 0.1251 & 0.1510  & 0.2284 & 0.3414
 & 0.1770  & 0.1922  & 0.1832 \\
PresRecST     & 0.2238 & 0.1749 & 0.1290  & 0.1512  & 0.2338 & 0.3465
& 0.1805  & 0.2001  & 0.1879 \\
SMRGAT      & 0.2461  & 0.1873  & 0.1343  & 0.1821 & 0.2777 & 0.3970
& 0.2093   & 0.2237  & 0.2007 \\
SCEIKG     & 0.2482  & 0.1942  & 0.1446  & 0.1722 & 0.2691 & 0.3973 
& 0.2033   & 0.2256  & 0.2119 \\
TCMRGCL    &\underline{0.2543} &\underline{0.1984} &\underline{0.1477}  &\underline{0.1852}  &\underline{0.2841} &\underline{0.4192}
& \underline{0.2143} & \underline{0.2337} & \underline{0.2184} \\
\textbf{FMASH\_RS (Ours)}  &\textbf{0.2629} & \textbf{0.2071} & \textbf{0.1527} &\textbf{0.1924} & \textbf{0.2967} & \textbf{0.4338}
& \textbf{0.2222} & \textbf{0.2439} & \textbf{0.2259} \\
\cline{1-10}

\%Improv. over TCMPR        & 43.43\%  & 43.12\%  & 32.67\%  & 43.80\%  & 39.49\%  & 29.46\%  & 43.63\%  & 41.64\%  & 31.87\% \\
\%Improv. over SMGCN        & 36.01\%  & 32.08\%  & 32.44\%  & 52.09\% & 46.08\%  & 43.31\%  & 45.32\%  & 37.80\%  & 35.27\%  \\
\%Improv. over KDHR         & 22.97\%  & 24.76\%  & 22.06\%  & 27.42\%  & 29.90\%  & 27.07\%  & 25.54\%  & 26.90\%  & 23.31\% \\
\%Improv. over PresRecST    & 17.47\%  & 18.41\%  & 18.37\%  & 27.25\%  & 26.90\%  & 25.19\%  & 23.10\%  & 21.89\%  & 20.22\% \\
\%Improv. over SMRGAT       & 6.83\%   & 10.57\%  & 13.70\%  & 5.66\%   & 6.84\%  & 9.27\%   & 6.16\%   & 9.03\%  & 12.56\% \\
\%Improv. over SCEIKG       & 5.92\%  & 6.64\%  & 5.60\%  & 11.73\%    & 10.26\%  & 9.19\%  & 9.30\%  & 8.11\%   & 6.61\% \\
\%Improv. over TCMRGCL       & 3.38\%  & 4.38\%  & 3.39\%  & 3.89\%    & 4.43\%    & 3.48\%   & 3.69\%   & 4.36\%   & 3.43\% \\
\hline
\end{tabular}
}
\caption{Comparison of FMASH\_RS with baseline models on Dataset1 under different evaluation metrics. The best and second-best scores in each metric are highlighted in \textbf{bold} and \underline{underline}, respectively.}
\label{tab:rs_result_data1}
\end{table*}

\begin{table*}[!t]
\centering
\small
\setlength{\tabcolsep}{4.5pt}
\renewcommand{\arraystretch}{1.12}
\resizebox{\textwidth}{!}{
\begin{tabular}{|l|ccc|ccc|ccc|}
\hline
\multirow{2}{*}{\textbf{Model}} 
& \multicolumn{3}{c|}{\textbf{Precision}}
& \multicolumn{3}{c|}{\textbf{Recall}}
& \multicolumn{3}{c|}{\textbf{F1-score}} \\
\cline{2-10}
& \textbf{P@5}  & \textbf{P@10}  & \textbf{P@20}  
& \textbf{R@5}  & \textbf{R@10}  & \textbf{R@20}
& \textbf{F1@5} & \textbf{F1@10} & \textbf{F1@20} \\
\hline

KDHR      & 0.2698   & 0.2103  & 0.1537  & 0.1932  & 0.2967  & 0.4299  & 0.2251  & 0.2461   & 0.2264 \\
SMRGAT    & 0.2710   & 0.2157  & 0.1574  & 0.1892  & 0.3009  & 0.4334   & 0.2228  & 0.2512   & 0.2309 \\
SCEIKG     & 0.2735  & 0.2189  & 0.1621  & 0.1921   & 0.3066  & 0.4482 & 0.2257  & 0.2555  & 0.2381 \\
PresRecST    & 0.2807  & 0.2248  & 0.1656  & 0.1988  & 0.3173 & 0.4618 & 0.2327  & 0.2632  & 0.2438 \\
SMGCN     & 0.2878   & 0.2287   & 0.1664   & 0.2070  & 0.3229  & 0.4631 & 0.2409  & 0.2677  & 0.2449 \\
TCMRGCL  & \underline{0.2884} & \underline{0.2310}  &\underline{0.1689} &\underline{0.2082} & \underline{0.3266} &\underline{0.4707}   &\underline{0.2418}  & \underline{0.2706}  & \underline{0.2486} \\
\textbf{FMASH\_RS (Ours)}  & \textbf{0.2960}  & \textbf{0.2352}  & \textbf{0.1722} & \textbf{0.2122} & \textbf{0.3335} & \textbf{0.4784}  &\textbf{0.2472} & \textbf{0.2759} & \textbf{0.2532} \\
\cline{1-10}
\%Improv. over SMGCN     & 2.85\% & 2.84\% & 3.49\%  & 2.51\%  & 3.28\% & 3.30\%  & 2.61\%  & 3.06\%  & 3.39\% \\
\%Improv. over KDHR      & 9.71\%   & 11.84\%  & 12.04\%   & 9.83\%      & 12.41\%  & 11.28\% & 9.82\%  & 12.11\%  & 11.84\% \\
\%Improv. over SMRGAT    & 9.23\%  & 9.04\%  & 9.40\%  & 12.16\%  & 10.83\%  & 10.38\%  & 10.95\%  & 9.83\%   & 9.66\% \\
\%Improv. over SCEIKG    & 8.23\%  & 7.45\%  & 6.23\% & 10.46\%  & 8.77\% & 6.74\%  & 9.53\% & 7.98\%  & 6.34\% \\
\%Improv. over PresRecST  & 5.45\% & 4.63\%  & 3.99\% & 6.74\% & 5.11\% & 3.59\%  & 6.23\%  & 4.82\%  & 3.86\% \\
\%Improv. over TCMRGCL   & 2.64\% & 1.82\%  & 1.95\%  & 1.92\% & 2.11\%  & 1.64\%  & 2.23\%  & 1.96\%  & 1.85\% \\
\hline
\end{tabular}
}
\caption{Comparison of FMASH\_RS with baseline models on Dataset2 under different evaluation metrics. The best and second-best scores in each metric are highlighted in \textbf{bold} and \underline{underline}, respectively.}
\label{tab:rs_result_data2}
\end{table*}

\begin{table*}[!t]
\centering
\renewcommand{\arraystretch}{1.2}
\setlength{\tabcolsep}{6pt}
\resizebox{\textwidth}{!}{
\begin{tabular}{|c|l|ccc|ccc|ccc|}
\hline
\multirow{2}{*}{\textbf{Type}} & \multirow{2}{*}{\textbf{Models}} 
& \multicolumn{3}{c|}{\textbf{Matched Precision}}
& \multicolumn{3}{c|}{\textbf{Matched Recall}}
& \multicolumn{3}{c|}{\textbf{Matched F1-score}} \\
\cline{3-11}
& & \textbf{MP@5} & \textbf{MP@10} & \textbf{MP@20}
& \textbf{MR@5} & \textbf{MR@10} & \textbf{MR@20}
& \textbf{MF1@5} & \textbf{MF1@10} & \textbf{MF1@20} \\
\hline
\multirow{4}{*}{\textbf{RS}}
& KDHR                  & 0.3444 & 0.2492 & 0.1839 & 0.2358 & 0.3156 & 0.4401 & 0.2799 & 0.2785 & 0.2594 \\
& PresRecST             & 0.3589 & 0.2723 & 0.1940 & 0.2506 & 0.3500 & 0.4675 & 0.2951 & 0.3063 & 0.2742 \\
& TCMRGCL               & 0.3785 & 0.3111 & 0.2163 & 0.2749 & 0.4065 & 0.5280 & 0.3185 & 0.3525 & 0.3069 \\
& \textbf{FMASH\_RS}    & \textbf{0.4022} & \textbf{0.3145} & \textbf{0.2318} & \textbf{0.2872} & \textbf{0.4089} & \textbf{0.5482} & \textbf{0.3351} & \textbf{0.3555} & \textbf{0.3258} \\
\hline
\multirow{2}{*}{\textbf{Seq}}
& Transformer\_baseline & 0.3582 & 0.2701 & 0.1457 & 0.2314 & 0.3219 & 0.3389 & 0.2632 & 0.2767 & 0.1936 \\
& \textbf{FMASH\_Seq}   & \textbf{0.4858} & \textbf{0.3581} & \textbf{0.1976} & \textbf{0.3370} & \textbf{0.4450} & \textbf{0.4697} & \textbf{0.3980} & \textbf{0.3969} & \textbf{0.2781} \\
\hline
\end{tabular}
}
\caption{Performance evaluation of two types of TCM formula recommendation model on Dataset1. RS denotes the traditional unordered herbal formula recommendation model, while Seq represents the herb sequential generative model.}
\label{tab:seq_result}
\end{table*}

\subsection{Ablation Study}
\subsubsection{Ablation Study of HGE, MLFI and MRM}
Based on the FMASH\_RS model, we have conducted an ablation experiment on three key functional modules and their different combinations on both datasets, as shown in Table \ref{tab:ablations}, to demonstrate their importance and effectiveness. The model with the corresponding key module HGE, MLFI, and MRM is denoted as subscripts 1, 2, and 3 in the Table, respectively. FMASH\_RS$_{base}$ represents a base model without any key functional modules, while the  "+" represents a combination of the corresponding functional modules. The findings from each dataset suggest that the model equipped with a single functional module can achieve an improvement in the Precision, Recall, and F1-score metrics. When the model is equipped with any combination of two functional modules, there can be significant improvements in every key metric compared to the base model. The most substantial improvement in performance is observed with the complete model: FMASH\_RS, i.e., the model with all key modules. Specifically, in Dataset1, Precision@5, Recall@5, and F1-score@5 increase by 12.06\%, 21.38\%, and 17.44\%, respectively, compared to the base model. In Dataset2, Precision@5, Recall@5, and F1-score@5 rise by 3.53\%, 4.84\%, and 4.22\%, respectively, compared to the base model. These results indicate that the effective integration and utilization of multiscale relations and features of symptoms and herbs provides the model with more granular information. This demonstrates the importance of the FMASH framework, which exhibits a superior ability to handle the complexity involved in formula recommendation and provide superior precision and quality in this task.

\begin{table*}[!t]
\centering
\resizebox{\textwidth}{!}{
\begin{tabular}{|l|c|ccc|ccc|ccc|}
\hline
\multirow{2}{*}{\textbf{Dataset}} 
&\multirow{2}{*}{\textbf{Model}} 
& \multicolumn{3}{c|}{\textbf{Precision}}
& \multicolumn{3}{c|}{\textbf{Recall}}
& \multicolumn{3}{c|}{\textbf{F1-score}} \\
\cline{3-11}
& 
& \textbf{P@5}  & \textbf{P@10}  & \textbf{P@20}  
& \textbf{R@5}  & \textbf{R@10}  & \textbf{R@20}
& \textbf{F1@5} & \textbf{F1@10} & \textbf{F1@20} \\
\hline
\multirow{9}{*}{Dataset1} 
&FMASH\_RS\textsubscript{base} & 0.2346 & 0.1853 & 0.1420  & 0.1585 & 0.2492  & 0.3804  & 0.1892 & 0.2125 & 0.2068 \\
&FMASH\_RS\textsubscript{1} & 0.2542 & 0.2028 & 0.1493 & 0.1837 &  0.2898  & 0.4222  &0.2132  & 0.2386 & 0.2206 \\
&FMASH\_RS\textsubscript{2} & 0.2586 & 0.2044  & 0.1520 & 0.1882 & 0.2924  & 0.4307  & 0.2178 & 0.2406 & 0.2247 \\
&FMASH\_RS\textsubscript{3} & 0.2577 & 0.2012 & 0.1495  & 0.1839 & 0.2846  & 0.4208  & 0.2146 & 0.2358 & 0.2207 \\
&FMASH\_RS\textsubscript{1+2} & 0.2599 & 0.2053 & 0.1520 & 0.1879 & 0.2924  & 0.4288  & 0.2181 & 0.2412 & 0.2244 \\
&FMASH\_RS\textsubscript{1+3} & 0.2611 & 0.2061 & 0.1515 & 0.1912 & 0.2960  & 0.4300  & 0.2207 & 0.2430 & 0.2241 \\
&FMASH\_RS\textsubscript{2+3} & 0.2610 & 0.2068 & \textbf{0.1532} & 0.1905 & 0.2961 & \textbf{0.4348} & 0.2202 & 0.2435  & \textbf{0.2265} \\
&\textbf{FMASH\_RS}  &\textbf{0.2629} & \textbf{0.2071} & 0.1527 &\textbf{0.1924} & \textbf{0.2967} & 0.4338
& \textbf{0.2222} & \textbf{0.2439} & 0.2259 \\ \hline
\multirow{9}{*}{Dataset2} 
&FMASH\_RS\textsubscript{base} & 0.2859 & 0.2288 & 0.1678 & 0.2024 & 0.3203 & 0.4636 & 0.2372 & 0.2669 &0.2464 \\
&FMASH\_RS\textsubscript{1} &0.2891 &0.2291 &0.1685
&0.2058 &0.3248 &0.4703 &0.2404 &0.2687 &0.2474 \\
&FMASH\_RS\textsubscript{2} &0.2889 &0.2278 &0.1686
&0.2055 &0.3222 &0.4687 &0.2402 &0.2669 &0.2475 \\
&FMASH\_RS\textsubscript{3} &0.2906 &0.2308 &0.1688 &0.2061 &0.3252 &0.4682 &0.2411 &0.2700 &0.2482 \\
&FMASH\_RS\textsubscript{1+2} &0.2942 &0.2344 &0.1720 &0.2104 &0.3321 &\textbf{0.4798} &0.2453 &0.2756 &0.2530 \\
&FMASH\_RS\textsubscript{1+3} &0.2929 &0.2340 &0.1710
&0.2088 &0.3319 &0.4778 &0.2438 &0.2745 &0.2516 \\
&FMASH\_RS\textsubscript{2+3} &0.2934 &0.2324 &0.1712
&0.2101 &0.3289 &0.4775  &0.2449 &0.2724 &0.2522 \\
&\textbf{FMASH\_RS}  & \textbf{0.2960}  & \textbf{0.2352}  & \textbf{0.1722} & \textbf{0.2122} & \textbf{0.3335} & 0.4784  &\textbf{0.2472} & \textbf{0.2759} & \textbf{0.2532} \\ \hline
\end{tabular}
}
\caption{Performance of different combinations of functional modules in FMASH\_RS compared to the base model on both two datasets, which demonstrates the importance and effectiveness of the key modules. The subscript 1, 2, and 3 represents the model with the module HGE, MLFI, and MRM, respectively, and "+" represents the combination of modules.}
\label{tab:ablations}
\end{table*}

\begin{figure*}[htbp]
\centering
\includegraphics[width=0.495\textwidth]{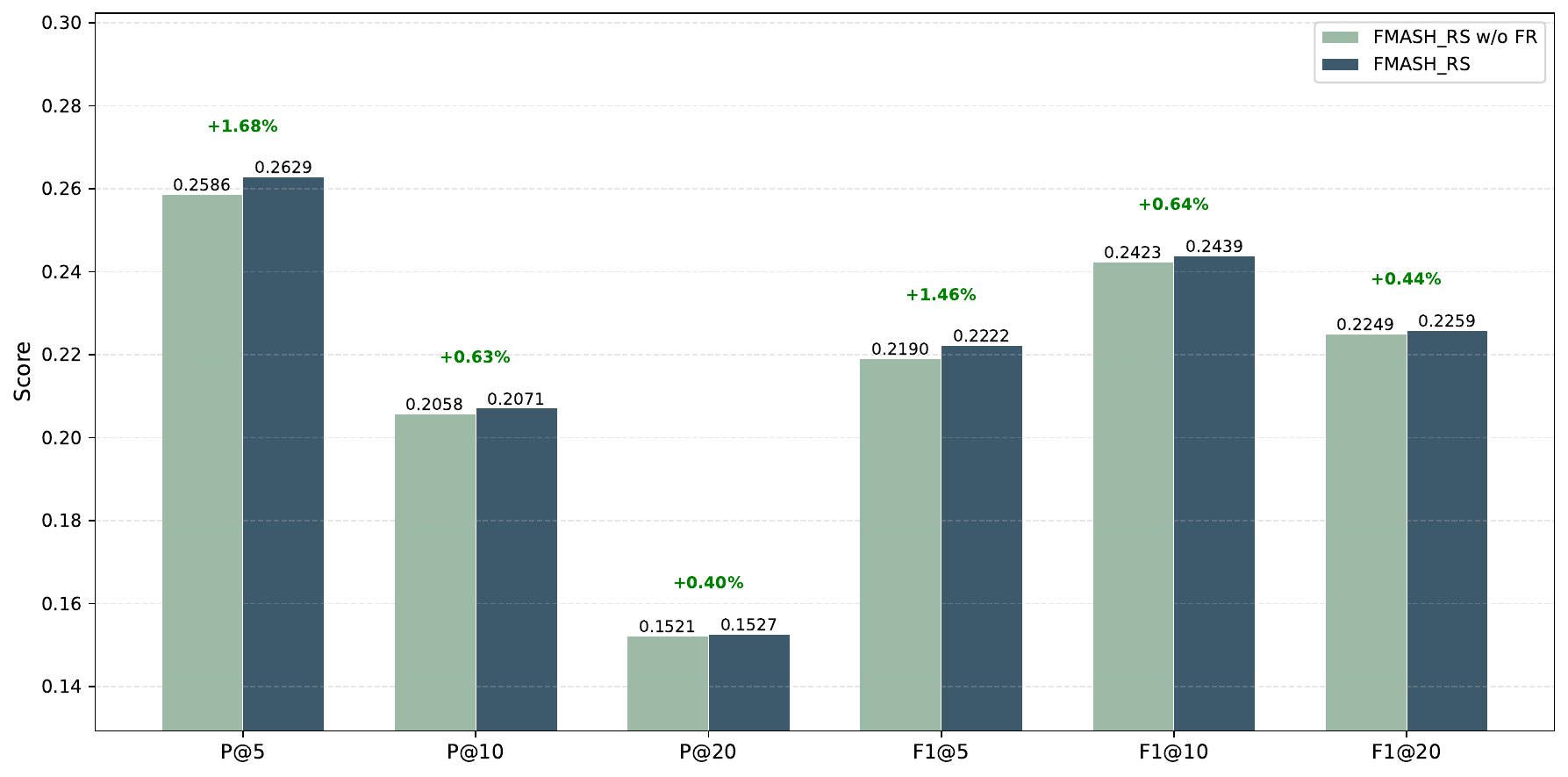}
\hfill
\includegraphics[width=0.495\textwidth]{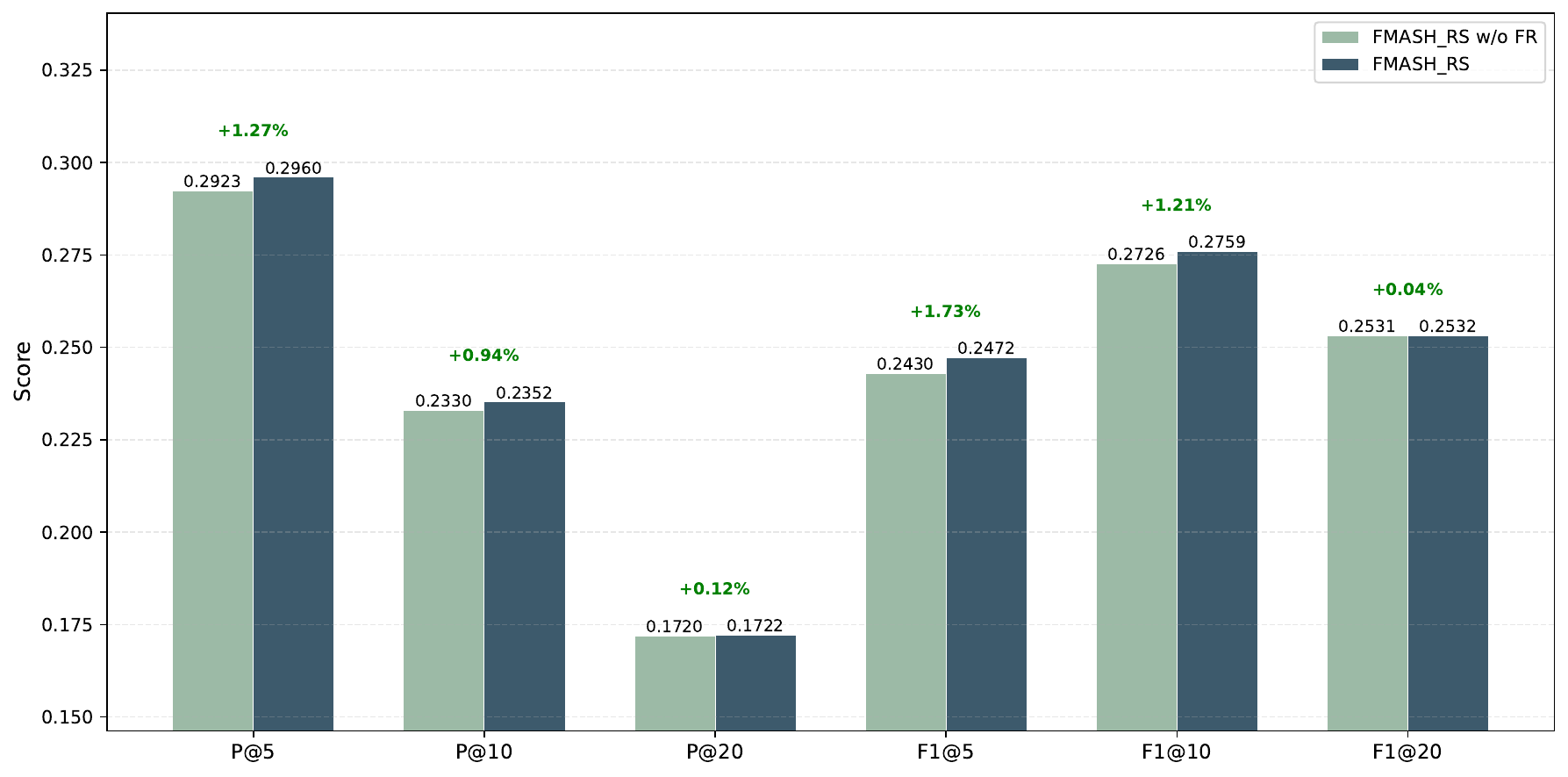}
\caption{Effect of Feature Refinement (FR) module on the two datasets. Each pair of bars compares FMASH\_RS w/o FR and the complete FMASH\_RS. Left: results on Dataset1. Right: results on Dataset2.}
\label{fig:performance_comparison}
\end{figure*}

\begin{table*}[hbp]
\centering
\small
\setlength{\tabcolsep}{6pt}
\renewcommand{\arraystretch}{1.5}
\resizebox{\textwidth}{!}{%
\begin{tabular}{|c|p{2.3cm}|p{4.0cm}|p{3.5cm}|p{3.5cm}|c|c|}
\hline
\multirow{2}{*}{\textbf{Case}} &
\multirow{2}{*}{\textbf{Symptoms}} &
\multicolumn{1}{c|}{\multirow{2}{*}{\textbf{FMASH\_RS Output}}} &
\multicolumn{1}{c|}{\multirow{2}{*}{\textbf{FMASH\_Seq Output}}} &
\multicolumn{1}{c|}{\multirow{2}{*}{\textbf{Ground Truth Formula}}} &
\multicolumn{2}{c|}{\textbf{P@5}} \\ \cline{6-7}
 & & & & & \textbf{FMASH\_RS} & \textbf{FMASH\_Seq} \\
\hline

\textbf{1} &
Head Sweat\begin{CJK}{UTF8}{gbsn}(头汗)\end{CJK},
Irritability\begin{CJK}{UTF8}{gbsn}(烦躁)\end{CJK} &
\textbf{Ru Xiang}\begin{CJK}{UTF8}{gbsn}(乳香)\end{CJK},
\textbf{Zhu Sha}\begin{CJK}{UTF8}{gbsn}(朱砂)\end{CJK},
\textbf{Fu Rong Ye}\begin{CJK}{UTF8}{gbsn}(芙蓉叶)\end{CJK},
\textbf{Wei Ling Xian}\begin{CJK}{UTF8}{gbsn}(威灵仙)\end{CJK},
Xiong Huang\begin{CJK}{UTF8}{gbsn}(雄黄)\end{CJK},
\zyc{Li Lu}\begin{CJK}{UTF8}{gbsn}(藜芦)\end{CJK},
Bai Fan\begin{CJK}{UTF8}{gbsn}(白矾)\end{CJK},
Jiang Can\begin{CJK}{UTF8}{gbsn}(僵蚕)\end{CJK},
Tu Mu Xiang\begin{CJK}{UTF8}{gbsn}(土木香)\end{CJK},
Hua Shi\begin{CJK}{UTF8}{gbsn}(滑石)\end{CJK},
Xiang Fu\begin{CJK}{UTF8}{gbsn}(香附)\end{CJK},
Chuan Xiong\begin{CJK}{UTF8}{gbsn}(川芎)\end{CJK},
Li\begin{CJK}{UTF8}{gbsn}(藜)\end{CJK},
Gan Song\begin{CJK}{UTF8}{gbsn}(甘松)\end{CJK},
\zyc{Xi Xin}\begin{CJK}{UTF8}{gbsn}(细辛)\end{CJK},
\zyc{Ku Shen}\begin{CJK}{UTF8}{gbsn}(苦参)\end{CJK},
Tian Ma\begin{CJK}{UTF8}{gbsn}(天麻)\end{CJK},
Yi Zhi Ren\begin{CJK}{UTF8}{gbsn}(益智仁)\end{CJK},
Da Suan\begin{CJK}{UTF8}{gbsn}(大蒜)\end{CJK},
Hu Jiao\begin{CJK}{UTF8}{gbsn}(胡椒)\end{CJK},
Shi Chang Pu\begin{CJK}{UTF8}{gbsn}(石菖蒲)\end{CJK} &
\textbf{Wei Ling Xian}\begin{CJK}{UTF8}{gbsn}(威灵仙)\end{CJK},
Ku Shen\begin{CJK}{UTF8}{gbsn}(苦参)\end{CJK},
\textbf{Zhu Sha}\begin{CJK}{UTF8}{gbsn}(朱砂)\end{CJK},
Bai Fan\begin{CJK}{UTF8}{gbsn}(白矾)\end{CJK},
\textbf{Ru Xiang}\begin{CJK}{UTF8}{gbsn}(乳香)\end{CJK},
Jiang Can\begin{CJK}{UTF8}{gbsn}(僵蚕)\end{CJK} &
\textbf{Wei Ling Xian}\begin{CJK}{UTF8}{gbsn}(威灵仙)\end{CJK},
\textbf{Di Long}\begin{CJK}{UTF8}{gbsn}(地龙)\end{CJK},
\textbf{Tian Zhu Huang}\begin{CJK}{UTF8}{gbsn}(天竺黄)\end{CJK},
\textbf{Da Suan}\begin{CJK}{UTF8}{gbsn}(大蒜)\end{CJK},
\textbf{Jiang Can}\begin{CJK}{UTF8}{gbsn}(僵蚕)\end{CJK},
\textbf{Xiong Huang}\begin{CJK}{UTF8}{gbsn}(雄黄)\end{CJK},
\textbf{Ru Xiang}\begin{CJK}{UTF8}{gbsn}(乳香)\end{CJK},
\textbf{Zhu Sha}\begin{CJK}{UTF8}{gbsn}(朱砂)\end{CJK},
\textbf{Fu Rong Ye}\begin{CJK}{UTF8}{gbsn}(芙蓉叶)\end{CJK},
\textbf{Peng Sha}\begin{CJK}{UTF8}{gbsn}(硼砂)\end{CJK} &
0.8 & 0.6 \\
\hline

\textbf{2} &
Dry Tongue\begin{CJK}{UTF8}{gbsn}(舌干)\end{CJK},
Lower Back Pain\begin{CJK}{UTF8}{gbsn}(腰痛)\end{CJK} &
\textbf{Wu Mei}\begin{CJK}{UTF8}{gbsn}(乌梅)\end{CJK},
\zyc{Yi Mu Cao}\begin{CJK}{UTF8}{gbsn}(益母草)\end{CJK},
Ma Yao\begin{CJK}{UTF8}{gbsn}(麻药)\end{CJK},
Long Chi\begin{CJK}{UTF8}{gbsn}(龙齿)\end{CJK},
Fu Man\begin{CJK}{UTF8}{gbsn}(腹满)\end{CJK},
Shan Yang Jiao\begin{CJK}{UTF8}{gbsn}(山羊角)\end{CJK},
Zhu Sha Gen\begin{CJK}{UTF8}{gbsn}(朱砂根)\end{CJK},
Ka Fei\begin{CJK}{UTF8}{gbsn}(咖啡)\end{CJK},
Ban Zhi Lian\begin{CJK}{UTF8}{gbsn}(半枝莲)\end{CJK},
\zyc{Ren Shen Lu}\begin{CJK}{UTF8}{gbsn}(人参芦)\end{CJK},
Ding Gong Teng\begin{CJK}{UTF8}{gbsn}(丁公藤)\end{CJK},
Mao Zhua Cao\begin{CJK}{UTF8}{gbsn}(猫爪草)\end{CJK},
Ci Wei Pi\begin{CJK}{UTF8}{gbsn}(刺猬皮)\end{CJK},
Di Feng Pi\begin{CJK}{UTF8}{gbsn}(地枫皮)\end{CJK},
Ye Ju Hua\begin{CJK}{UTF8}{gbsn}(野菊花)\end{CJK},
Bai Dou Kou\begin{CJK}{UTF8}{gbsn}(白豆蔻)\end{CJK},
Mo Han Lian\begin{CJK}{UTF8}{gbsn}(墨旱莲)\end{CJK},
Mi Hou Tao\begin{CJK}{UTF8}{gbsn}(猕猴桃)\end{CJK},
Ji Gu Cao\begin{CJK}{UTF8}{gbsn}(鸡骨草)\end{CJK},
Shi Chang Pu\begin{CJK}{UTF8}{gbsn}(石菖蒲)\end{CJK} &
\textbf{Fan Xie Ye}\begin{CJK}{UTF8}{gbsn}(番泻叶)\end{CJK},
\textbf{Chuan Bei Mu}\begin{CJK}{UTF8}{gbsn}(川贝母)\end{CJK},
\textbf{Xi Xin}\begin{CJK}{UTF8}{gbsn}(细辛)\end{CJK},
\textbf{Wu Mei}\begin{CJK}{UTF8}{gbsn}(乌梅)\end{CJK},
\textbf{Yi Zhi Ren}\begin{CJK}{UTF8}{gbsn}(益智仁)\end{CJK} &
\textbf{Wu Mei}\begin{CJK}{UTF8}{gbsn}(乌梅)\end{CJK},
\textbf{Jin Yin Hua}\begin{CJK}{UTF8}{gbsn}(金银花)\end{CJK},
\textbf{Chuan Bei Mu}\begin{CJK}{UTF8}{gbsn}(川贝母)\end{CJK},
\textbf{Fan Xie Ye}\begin{CJK}{UTF8}{gbsn}(番泻叶)\end{CJK},
\textbf{Hu Jiao}\begin{CJK}{UTF8}{gbsn}(胡椒)\end{CJK},
\textbf{Mu Xiang}\begin{CJK}{UTF8}{gbsn}(木香)\end{CJK},
\textbf{Xi Xin}\begin{CJK}{UTF8}{gbsn}(细辛)\end{CJK},
\textbf{Yi Zhi Ren}\begin{CJK}{UTF8}{gbsn}(益智仁)\end{CJK} &
0.2 & 1.0 \\
\hline
\end{tabular}%
}
\caption{Performance of the two types of TCM formula generation models: FMASH\_RS and FMASH\_Seq, under the same input of symptoms. The correct herbs in top-5 predictions by each model are in highlighted in bold, and the negative results are marked in red.}
\label{tab:herb_case_study_seq}
\end{table*}

\subsubsection{Ablation Study of FR}

Ablation analysis of the Feature Refinement (FR) module has been conducted on both datasets. In this study, the FMASH\_RS model without FR module (denoted as FMASH\_RS w/o FR) is compared with the complete FMASH\_RS model, and their performance has been evaluated according to key metrics: Precision and F1-score. As shown in Figure \ref{fig:performance_comparison}, in terms of Precision and the F1-score, the complete FMASH\_RS model that incorporates the FR module consistently outperforms the model without FR. Specifically, in Dataset1, the complete model exhibits an improvement of 1.68\% and 1.46\%, respectively, in terms of Precision@5 and F1-score@5. Similarly, in Dataset2, the complete model achieves improvements of 1.27\% and 1.73\% in the corresponding metrics, respectively. These results demonstrate the effectiveness of FR, indicating that incorporating the FR module can effectively preserve the original characteristics during feature compression and enhance feature integration.

\subsection{Effect of MLFI}
\begin{figure}[ht] 
  \centering
  \includegraphics[width=1\columnwidth]{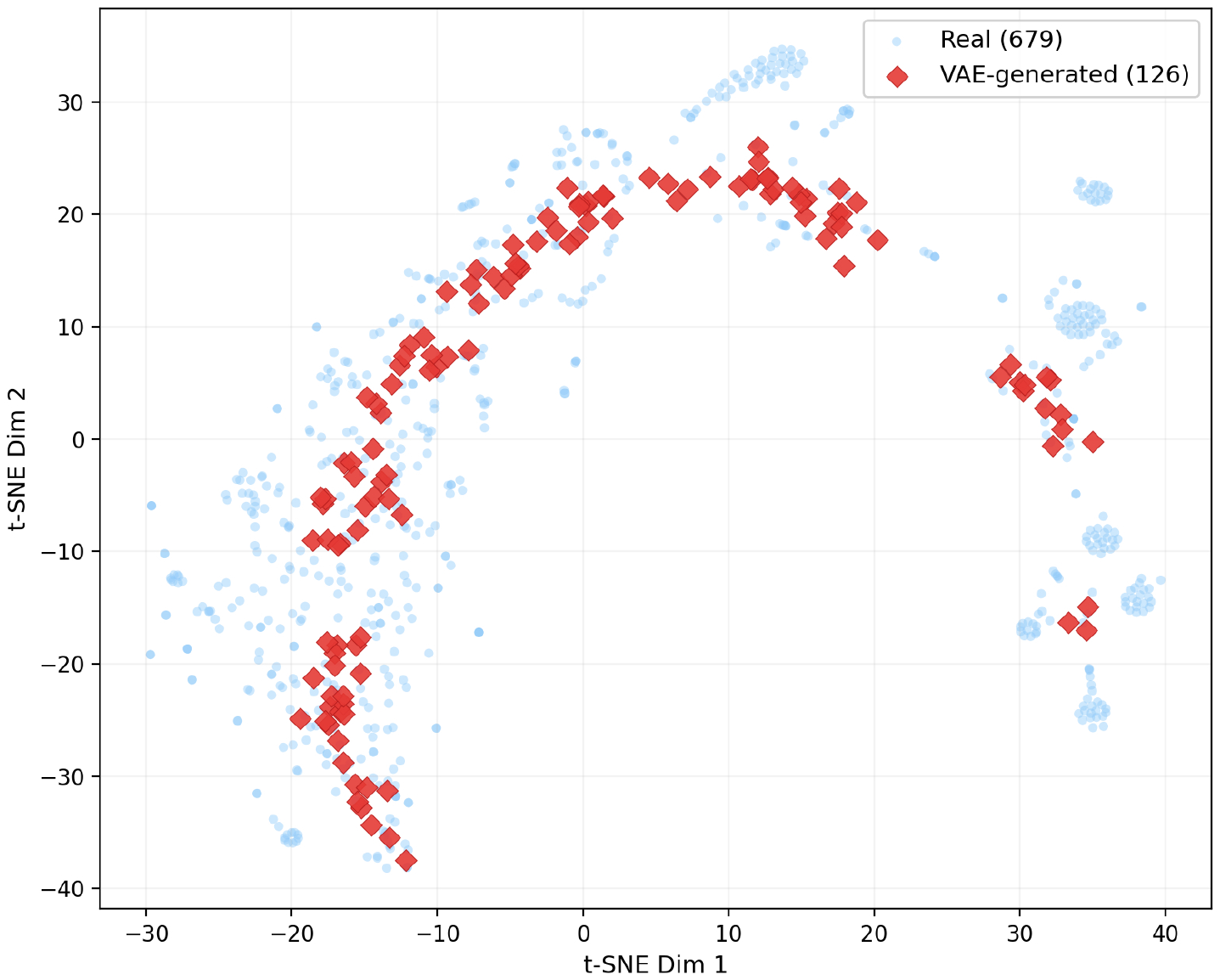} 
  \caption{The visualized distributions of molecular-level representation embeddings of herbs generated by MLFI module.}
  \label{fig: vae_tsne_generate}
\end{figure}

To evaluate the performance of the developed Molecular-Level Feature Integration (MLFI) module more vividly and thoroughly, for Dataset1, we visualize the molecular-level representation of herbs using t-SNE as illustrated in Figure \ref{fig: vae_tsne_generate}. We map all molecular-level representations of 805 herbs in Dataset1 into a two-dimensional space. Of these, the 679 herbs for which molecular information was available were used as real samples to directly obtain their molecular-level representations, while the remaining 126 herbs for which molecular information was missing had their molecular-level representations generated by our proposed VAE-like method in the MLFI module. 

As shown in Figure \ref{fig: vae_tsne_generate}, the molecular-level representations of herbs with missing molecular information generated by MLFI manifest a distribution pattern that is highly consistent with that of herbs with known molecular information. Most of the VAE-generated samples are distributed in regions with relatively high densities of real samples, and there is a significant overlap between their distributions, indicating that the supplemented molecular information is semantically aligned with the original molecular space of herbs. This observation suggests that the proposed MLFI can successfully bridge the gap caused by incomplete molecular-level data and enhance the completeness and consistency of herb embeddings, which further benefit the downstream prediction task.

Furthermore, the embedding distributions of molecular-level representations of herbs generated by the MLFI module are relatively compact, forming clear clusters with obvious distinctions. This suggests that the module is capable of grouping similar herbs together while distinguishing different molecular-level features of herbs, which improves the model's performance in recommending TCM formulas.

\subsection{Case Study of FMASH}

\begin{table*}[htbp]
\centering
\small
\setlength{\tabcolsep}{3pt}
\renewcommand{\arraystretch}{1.0}
\scalebox{0.85}{%
\begin{tabular}{|c|p{1.9cm}|c|c|}
\hline
\textbf{Case} & \textbf{Symptoms} & \textbf{Model Output vs.\ Ground Truth} & \textbf{Evaluate Metrics} \\ \hline
1 & Tearing\begin{CJK}{UTF8}{gbsn}(流泪)\end{CJK} &
\begin{tabular}[c]{@{}p{6cm}@{}}
\textbf{Generated TCM Formula:}\\
\textbf{Xi Xin}\begin{CJK}{UTF8}{gbsn}(细辛)\end{CJK},
\textbf{Hua Shi}\begin{CJK}{UTF8}{gbsn}(滑石)\end{CJK},
\textbf{Dan Zhu Ye}\begin{CJK}{UTF8}{gbsn}(淡竹叶)\end{CJK},
\textbf{Quan Xie}\begin{CJK}{UTF8}{gbsn}(全蝎)\end{CJK},
\textbf{Mu Xiang}\begin{CJK}{UTF8}{gbsn}(木香)\end{CJK},
\textbf{E Zhu}\begin{CJK}{UTF8}{gbsn}(莪术)\end{CJK},
Chuan Xiong\begin{CJK}{UTF8}{gbsn}(川芎)\end{CJK},
Wu Tou\begin{CJK}{UTF8}{gbsn}(乌头)\end{CJK},
\textbf{Yu Gan Zi}\begin{CJK}{UTF8}{gbsn}(余甘子)\end{CJK},
\textbf{Qing Mu Xiang}\begin{CJK}{UTF8}{gbsn}(青木香)\end{CJK},
\textbf{Fang Ji}\begin{CJK}{UTF8}{gbsn}(防己)\end{CJK},
\textbf{Li}\begin{CJK}{UTF8}{gbsn}(李)\end{CJK},
\textbf{Gan Qi}\begin{CJK}{UTF8}{gbsn}(干漆)\end{CJK},
\textbf{Lv Dou}\begin{CJK}{UTF8}{gbsn}(绿豆)\end{CJK},
\textbf{Tao}\begin{CJK}{UTF8}{gbsn}(桃)\end{CJK},
Long Dan\begin{CJK}{UTF8}{gbsn}(龙胆)\end{CJK},
\textbf{Yong Sui Ke}\begin{CJK}{UTF8}{gbsn}(雍穗壳)\end{CJK},
Tan Xiang\begin{CJK}{UTF8}{gbsn}(檀香)\end{CJK},
Mu Dan\begin{CJK}{UTF8}{gbsn}(牡丹)\end{CJK},
\textbf{Ji Guan Hua}\begin{CJK}{UTF8}{gbsn}(鸡冠花)\end{CJK}\\[4pt]
\textbf{Ground Truth TCM Formula:}\\
Che Qian Zi\begin{CJK}{UTF8}{gbsn}(车前子)\end{CJK},
Teng Huang\begin{CJK}{UTF8}{gbsn}(藤黄)\end{CJK},
Shi Chang Pu\begin{CJK}{UTF8}{gbsn}(石菖蒲)\end{CJK},
Xiong Dan\begin{CJK}{UTF8}{gbsn}(熊胆)\end{CJK},
Gao Liang Jiang\begin{CJK}{UTF8}{gbsn}(高良姜)\end{CJK},
Tu Mu Xiang\begin{CJK}{UTF8}{gbsn}(土木香)\end{CJK},
Hua Shi\begin{CJK}{UTF8}{gbsn}(滑石)\end{CJK},
Gan Qi\begin{CJK}{UTF8}{gbsn}(干漆)\end{CJK},
Hu Po\begin{CJK}{UTF8}{gbsn}(琥珀)\end{CJK},
E Zhu\begin{CJK}{UTF8}{gbsn}(莪术)\end{CJK},
Gua Ti\begin{CJK}{UTF8}{gbsn}(瓜蒂)\end{CJK},
Li\begin{CJK}{UTF8}{gbsn}(李)\end{CJK},
Xiao Hui Xiang\begin{CJK}{UTF8}{gbsn}(小茴香)\end{CJK},
Chang Shan\begin{CJK}{UTF8}{gbsn}(常山)\end{CJK},
Qing Mu Xiang\begin{CJK}{UTF8}{gbsn}(青木香)\end{CJK},
Tian Ma\begin{CJK}{UTF8}{gbsn}(天麻)\end{CJK},
Mu Xiang\begin{CJK}{UTF8}{gbsn}(木香)\end{CJK},
Yu Gan Zi\begin{CJK}{UTF8}{gbsn}(余甘子)\end{CJK},
Yong Sui Ke\begin{CJK}{UTF8}{gbsn}(雍穗壳)\end{CJK},
Lu Gan Shi\begin{CJK}{UTF8}{gbsn}(炉甘石)\end{CJK},
Fang Ji\begin{CJK}{UTF8}{gbsn}(防己)\end{CJK},
Xi Xin\begin{CJK}{UTF8}{gbsn}(细辛)\end{CJK},
Ji Guan Hua\begin{CJK}{UTF8}{gbsn}(鸡冠花)\end{CJK},
Dan Zhu Ye\begin{CJK}{UTF8}{gbsn}(淡竹叶)\end{CJK},
Qi She\begin{CJK}{UTF8}{gbsn}(蕲蛇)\end{CJK},
Tao\begin{CJK}{UTF8}{gbsn}(桃)\end{CJK},
Xuan Ming Fen\begin{CJK}{UTF8}{gbsn}(玄明粉)\end{CJK},
Quan Xie\begin{CJK}{UTF8}{gbsn}(全蝎)\end{CJK},
Gao Ben\begin{CJK}{UTF8}{gbsn}(藁本)\end{CJK},
Shan Zha\begin{CJK}{UTF8}{gbsn}(山楂)\end{CJK},
She Chuan Zi\begin{CJK}{UTF8}{gbsn}(蛇床子)\end{CJK},
Lv Dou\begin{CJK}{UTF8}{gbsn}(绿豆)\end{CJK},
Ban Miao\begin{CJK}{UTF8}{gbsn}(斑蝥)\end{CJK},
Qing Hao\begin{CJK}{UTF8}{gbsn}(青蒿)\end{CJK}
\end{tabular} &
\begin{tabular}[c]{@{}c@{}}
P@5=1.0\\P@10=0.8\\P@20=0.75
\end{tabular} \\ \hline

2 & Leukorrhea\begin{CJK}{UTF8}{gbsn}(白带)\end{CJK}, Gan Ji\begin{CJK}{UTF8}{gbsn}(疳积)\end{CJK} &
\begin{tabular}[c]{@{}p{6cm}@{}}
\textbf{Generated TCM Formula:}\\
\textbf{Fan Xie Ye}\begin{CJK}{UTF8}{gbsn}(番泻叶)\end{CJK},
\textbf{Xi Xin}\begin{CJK}{UTF8}{gbsn}(细辛)\end{CJK},
\textbf{Zhi Mu}\begin{CJK}{UTF8}{gbsn}(知母)\end{CJK},
\textbf{Yi Zhi Ren}\begin{CJK}{UTF8}{gbsn}(益智仁)\end{CJK},
\textbf{Mai Dong}\begin{CJK}{UTF8}{gbsn}(麦冬)\end{CJK},
\textbf{Bai Zhu}\begin{CJK}{UTF8}{gbsn}(白术)\end{CJK},
\textbf{Ren Shen}\begin{CJK}{UTF8}{gbsn}(人参)\end{CJK},
\textbf{Qian Niu}\begin{CJK}{UTF8}{gbsn}(牵牛)\end{CJK},
Gua Ti\begin{CJK}{UTF8}{gbsn}(瓜蒂)\end{CJK},
\textbf{Xing}\begin{CJK}{UTF8}{gbsn}(杏)\end{CJK},
Mu Xiang\begin{CJK}{UTF8}{gbsn}(木香)\end{CJK},
\textbf{Huang Lian}\begin{CJK}{UTF8}{gbsn}(黄连)\end{CJK},
Mu Dan\begin{CJK}{UTF8}{gbsn}(牡丹)\end{CJK},
Li\begin{CJK}{UTF8}{gbsn}(李)\end{CJK},
\textbf{Pu Tao}\begin{CJK}{UTF8}{gbsn}(葡萄)\end{CJK},
Jing Jie\begin{CJK}{UTF8}{gbsn}(荆芥)\end{CJK},
Chuan Xiong\begin{CJK}{UTF8}{gbsn}(川芎)\end{CJK},
E Zhu\begin{CJK}{UTF8}{gbsn}(莪术)\end{CJK},
\textbf{Hu Po}\begin{CJK}{UTF8}{gbsn}(琥珀)\end{CJK}\\[4pt]
\textbf{Ground Truth TCM Formula:}\\
Bi Bo\begin{CJK}{UTF8}{gbsn}(荜茇)\end{CJK},
Hua Jiao\begin{CJK}{UTF8}{gbsn}(花椒)\end{CJK},
Chen Xiang\begin{CJK}{UTF8}{gbsn}(沉香)\end{CJK},
Pu Tao\begin{CJK}{UTF8}{gbsn}(葡萄)\end{CJK},
Xing\begin{CJK}{UTF8}{gbsn}(杏)\end{CJK},
Quan Shen\begin{CJK}{UTF8}{gbsn}(全参)\end{CJK},
Zhi Mu\begin{CJK}{UTF8}{gbsn}(知母)\end{CJK},
Mai Dong\begin{CJK}{UTF8}{gbsn}(麦冬)\end{CJK},
Bai Fu Zi\begin{CJK}{UTF8}{gbsn}(白附子)\end{CJK},
Qian Niu\begin{CJK}{UTF8}{gbsn}(牵牛)\end{CJK},
Shen Qu\begin{CJK}{UTF8}{gbsn}(神曲)\end{CJK},
Huang Lian\begin{CJK}{UTF8}{gbsn}(黄连)\end{CJK},
Ren Shen\begin{CJK}{UTF8}{gbsn}(人参)\end{CJK},
Fan Xie Ye\begin{CJK}{UTF8}{gbsn}(番泻叶)\end{CJK},
Qing Pi\begin{CJK}{UTF8}{gbsn}(青皮)\end{CJK},
Bai Cao Shuang\begin{CJK}{UTF8}{gbsn}(百草霜)\end{CJK},
Bai Zhu\begin{CJK}{UTF8}{gbsn}(白术)\end{CJK},
Xi Xin\begin{CJK}{UTF8}{gbsn}(细辛)\end{CJK},
Yi Zhi Ren\begin{CJK}{UTF8}{gbsn}(益智仁)\end{CJK}
\end{tabular} &
\begin{tabular}[c]{@{}c@{}}
P@5=1.0\\P@10=0.9\\P@20=0.55
\end{tabular} \\ \hline

3 & Infertility\begin{CJK}{UTF8}{gbsn}(不孕)\end{CJK}, Epilepsy\begin{CJK}{UTF8}{gbsn}(癫痫)\end{CJK} &
\begin{tabular}[c]{@{}p{6cm}@{}}
\textbf{Generated TCM Formula:}\\
\textbf{Hua Shi}\begin{CJK}{UTF8}{gbsn}(滑石)\end{CJK},
\textbf{Gan Qi}\begin{CJK}{UTF8}{gbsn}(干漆)\end{CJK},
\textbf{Xi Xin}\begin{CJK}{UTF8}{gbsn}(细辛)\end{CJK},
\textbf{Gao Ben}\begin{CJK}{UTF8}{gbsn}(藁本)\end{CJK},
\textbf{Li}\begin{CJK}{UTF8}{gbsn}(李)\end{CJK},
\textbf{Bai Fan}\begin{CJK}{UTF8}{gbsn}(白矾)\end{CJK},
\textbf{Fan Xie Ye}\begin{CJK}{UTF8}{gbsn}(番泻叶)\end{CJK},
\textbf{Chuan Xiong}\begin{CJK}{UTF8}{gbsn}(川芎)\end{CJK},
\textbf{Tian Ma}\begin{CJK}{UTF8}{gbsn}(天麻)\end{CJK},
\textbf{Yi Zhi Ren}\begin{CJK}{UTF8}{gbsn}(益智仁)\end{CJK},
\textbf{Li Lu}\begin{CJK}{UTF8}{gbsn}(藜芦)\end{CJK},
\textbf{Tu Mu Xiang}\begin{CJK}{UTF8}{gbsn}(土木香)\end{CJK},
\textbf{Mu Dan}\begin{CJK}{UTF8}{gbsn}(牡丹)\end{CJK},
Fang Feng\begin{CJK}{UTF8}{gbsn}(防风)\end{CJK},
\textbf{Zhi Mu}\begin{CJK}{UTF8}{gbsn}(知母)\end{CJK},
Wei Ling Xian\begin{CJK}{UTF8}{gbsn}(威灵仙)\end{CJK},
Zhu Sha\begin{CJK}{UTF8}{gbsn}(朱砂)\end{CJK},
Ru Xiang\begin{CJK}{UTF8}{gbsn}(乳香)\end{CJK},
\textbf{Gua Ti}\begin{CJK}{UTF8}{gbsn}(瓜蒂)\end{CJK},
Tao\begin{CJK}{UTF8}{gbsn}(桃)\end{CJK}\\[4pt]
\textbf{Ground Truth TCM Formula:}\\
Mai Men Dong\begin{CJK}{UTF8}{gbsn}(麦门冬)\end{CJK},
Long Gu\begin{CJK}{UTF8}{gbsn}(龙骨)\end{CJK},
Shi Chang Pu\begin{CJK}{UTF8}{gbsn}(石菖蒲)\end{CJK},
Xiong Dan\begin{CJK}{UTF8}{gbsn}(熊胆)\end{CJK},
Jin Yin Hua\begin{CJK}{UTF8}{gbsn}(金银花)\end{CJK},
Dong Gua Zi\begin{CJK}{UTF8}{gbsn}(冬瓜子)\end{CJK},
Gao Liang Jiang\begin{CJK}{UTF8}{gbsn}(高良姜)\end{CJK},
Tu Mu Xiang\begin{CJK}{UTF8}{gbsn}(土木香)\end{CJK},
Hua Shi\begin{CJK}{UTF8}{gbsn}(滑石)\end{CJK},
Gan Qi\begin{CJK}{UTF8}{gbsn}(干漆)\end{CJK},
Zhi Mu\begin{CJK}{UTF8}{gbsn}(知母)\end{CJK},
Mai Dong\begin{CJK}{UTF8}{gbsn}(麦冬)\end{CJK},
Mu Dan\begin{CJK}{UTF8}{gbsn}(牡丹)\end{CJK},
E Zhu\begin{CJK}{UTF8}{gbsn}(莪术)\end{CJK},
Qian Dan\begin{CJK}{UTF8}{gbsn}(铅丹)\end{CJK},
Qian Niu\begin{CJK}{UTF8}{gbsn}(牵牛)\end{CJK},
Gua Ti\begin{CJK}{UTF8}{gbsn}(瓜蒂)\end{CJK},
Li\begin{CJK}{UTF8}{gbsn}(李)\end{CJK},
Fan Xie Ye\begin{CJK}{UTF8}{gbsn}(番泻叶)\end{CJK},
Tian Ma\begin{CJK}{UTF8}{gbsn}(天麻)\end{CJK},
Li Lu\begin{CJK}{UTF8}{gbsn}(藜芦)\end{CJK},
Bai Fan\begin{CJK}{UTF8}{gbsn}(白矾)\end{CJK},
Mu Xiang\begin{CJK}{UTF8}{gbsn}(木香)\end{CJK},
Lu Gan Shi\begin{CJK}{UTF8}{gbsn}(炉甘石)\end{CJK},
Dang Shen\begin{CJK}{UTF8}{gbsn}(党参)\end{CJK},
Bai Zhu\begin{CJK}{UTF8}{gbsn}(白术)\end{CJK},
Xi Xin\begin{CJK}{UTF8}{gbsn}(细辛)\end{CJK},
Yi Zhi Ren\begin{CJK}{UTF8}{gbsn}(益智仁)\end{CJK},
He Zi\begin{CJK}{UTF8}{gbsn}(诃子)\end{CJK},
Chuan Xiong\begin{CJK}{UTF8}{gbsn}(川芎)\end{CJK},
Ji Guan Hua\begin{CJK}{UTF8}{gbsn}(鸡冠花)\end{CJK},
Dan Zhu Ye\begin{CJK}{UTF8}{gbsn}(淡竹叶)\end{CJK},
Xuan Ming Fen\begin{CJK}{UTF8}{gbsn}(玄明粉)\end{CJK},
Gao Ben\begin{CJK}{UTF8}{gbsn}(藁本)\end{CJK},
Wu Tou\begin{CJK}{UTF8}{gbsn}(乌头)\end{CJK},
Che Qian\begin{CJK}{UTF8}{gbsn}(车前)\end{CJK}
\end{tabular} &
\begin{tabular}[c]{@{}c@{}}
P@5=1.0\\P@10=1.0\\P@20=0.75
\end{tabular} \\ \hline
\end{tabular}  %
}
\caption{Performance of the FMASH\_RS model with some typical symptoms as input. The correct herbs in top-5 predictions by the model are in highlighted in bold, and the negative results are marked in red.}
\label{tab:herb_case_study_rs}
\end{table*}

\subsubsection{Case Study of FMASH\_RS}
We have selected some typical cases from the Dataset1 to demonstrate the powerful performance of the FMASH framework in the TCM formula recommendation task, seeing the results in Table \ref{tab:herb_case_study_rs}. In Case 1 (Symptom: Tearing), the model achieved precision with Precision@5=1.0 and Precision@10 =0.8, which is clinically significant. For instance, the model's top-ranked recommendations included Xi Xin, which is known in TCM for dispelling wind-cold and alleviating pain that can manifest in the eyes, and Dan Zhu Ye, which effectively clears heat. This suggests that the model correctly identifies that "tearing" can be associated with both wind and heat pathologies and  provides relevant therapeutic herbs.

In Case 2 (Symptoms: Leukorrhea, Gan Ji), the model shows high precision (Precision@5 =1.0, Precision@10=0.9) in a more complex scenario. The analysis of the result indicates the model's deeper understanding of the TCM principles . The model not only suggested herbs like Yi Zhi Ren, which directly addresses leukorrhea by astringing essence, but also recommended the classic combination of Ren Shen and Bai Zhu. This pair is fundamental for tonifying the Spleen's Qi, targeting the root cause of many deficiency-pattern leukorrheas. 

In the highly complex Case 3 (Symptoms: Infertility, Epilepsy), the model also achieves remarkable precision at both Precision@5  and Precision@10. This case involves two disparate and challenging issues, while the model successfully addresses these by recommending classic herbs for extinguishing internal wind, such as Tian Ma and Gao Ben, which are used to control seizures and epilepsy. It also proposes Yi Zhi Ren, which nourishes the essence of the kidney and is a key strategy in the treatment of certain types of infertility. 

These results validate that the model has the ability to generate a coherent set of herbs that address multiple complex disease patterns and capture the underlying diagnostic and therapeutic logic of TCM. It has demonstrated significant practical potential for the FMASH framework in some typical cases.


\subsubsection{Case Study of FMASH\_Seq} 

Some typical cases in Dataset1 have been explored to compare the difference between the two types of TCM formula generation models: FMASH\_RS and FMASH\_Seq, seeing the results in Table \ref{tab:herb_case_study_seq}. In Case 1, the FMASH\_RS model yields a marginally higher P@5 than the FMASH\_Seq model (0.8 versus 0.6). However, its output is a fixed-length list (20 herbs in this case), where clinically irrelevant or contraindicated herbs may be inevitably included. By contrast, the FMASH\_Seq model is equipped with an end-of-sequence (EOS) token, and once the model judged the complete of the formula, it emits the EOS token and terminates the generation. Therefore, sequence-to-sequence models offer an explicit, data-driven stopping rule, whereas traditional TCM formula recommendation models are incapable of providing a principled-based criterion for truncation during the application.

In Case 2, the FMASH\_Seq model achieves a score of 1.0 in Precision@5, which is a significant improvement over FMASH\_RS (0.2 in Precision@5). This significant difference shows that the sequential model can identify the exact therapeutic subset of the pattern of “dry tongue \& lumbar pain”.  

\textbf{Incompatibility and Toxicity Profile:} In these cases, the FMASH\_RS recommendation routinely concatenates 20 herbs, among which \textit{Veratri nigri radix et rhizoma} (Li Lu), \textit{Asari radix et rhizoma} (Xi Xin) and \textit{Sophorae flavescentis radix} (Ku Shen) appear simultaneously—an ensemble that violates the classical incompatibility rules and may increase the toxic risk in the recommended formula. This defect arises because FMASH\_RS aggregates heterogeneous historical prescriptions with the same symptomatic label and ranks constituents by marginal frequency. This cross-formula pooling dissolves the original therapeutic contexts and their contraindication safeguards, resulting in pairs of antagonistic or toxic herbs to coexist. FMASH\_Seq is trained to generate a single coherent prescription and emit an end-of-sequence (EOS) signal upon completion. Therefore, FMASH\_Seq intrinsically preserves intra-formula compatibility and does not merge discordant TCM recipes.

\section{Summary and Discussion}
\label{sec:summary}
This study proposes a novel framework, FMASH, for TCM formula recommendation. FMASH systematically integrates the characteristics and relations of symptoms and herbs on different scales and provides effective refined representation embeddings, establishing an effective paradigm for building TCM formula recommendation models. This framework effectively integrates the molecular-level features of herbs with their macroscopic properties and captures the hierarchical local and global relations in heterogeneous graph networks of symptoms and herbs, providing effective unified embeddings of their multiscale features. Based on elaborately designed functional modules, FMASH has demonstrated its effectiveness through two different ways of generating TCM formulas. The experimental results show that the FMASH$\_$RS model outperforms existing baseline models in two commonly used datasets in terms of Precision, Recall, and F1-Score. In Dataset1, the average improvement compared to the current state-of-the-art (SOTA) model is 3.38\% in Precision@5, 3.89\% in Recall@5, and 3.69\% in F1-score@5. In Dataset2, Precision@5, Recall@5, and F1-score@5 increase by 2.64\%, 1.92\%, and 2.23\%, respectively, compared to the SOTA model. Moreover, the FMASH$\_$Seq model also exhibits significantly superior performance compared to the existing SOTA model, achieving relative improvements of 28.35\% in MP@5, 22.59\% in MR@5, and 24.96\% in MF1@5.

Comprehensive ablation studies have been conducted to demonstrate the effectiveness of the overall framework and the key functional modules developed, validating that the functional module-based framework effectively improves the model's performance in recommending TCM formulas. The case study demonstrates that, according to the FMASH framework, the model can generate a coherent set of herbs that captures the underlying diagnostic and therapeutic logic of TCM. Meanwhile, the FMASH$\_$Seq model has been found to generate formulas of effective length, which intrinsically preserves intra-formula compatibility and does not merge discordant recipes. These findings validate the effectiveness of our proposed framework and provide two distinct options in addressing the task of automatic recommendations of the TCM formula. 

Although the average performance of the model remains to be enhanced to align with the precision criteria of practical clinical applications, this study has substantiated the considerable promise of the FMASH framework in improving the effectiveness of TCM treatment. It provides personalized prescription suggestions and reduces dependence on the experience of the individual physician, serving as a valuable decision-support tool and offering a direction for the development of AI-based intelligent TCM formula recommendation systems. The current FMASH framework has not considered the dosage of herbs and the progress of a patient’s condition, which remains the direction for future improvement of the model.

\section*{Acknowledgments}
This work was supported by the National Natural Science Foundation of China under grant No.12301549. 

\section*{Data Availability }
The code and data related to this study will be made available upon request.

\bibliography{references}
\bibliographystyle{elsarticle-num}

\end{document}